\pdfoutput=1

\documentclass{article}
\usepackage[final,nonatbib]{nips_2016}

\usepackage[utf8]{inputenc} 
\usepackage[T1]{fontenc}    
\usepackage{hyperref}       
\usepackage{url}            
\usepackage{booktabs}       
\usepackage{amsfonts}       
\usepackage{nicefrac}       
\usepackage{microtype}      
\usepackage{algorithm}
\usepackage{algorithmic}
\usepackage{wrapfig}
\usepackage{enumitem}
\usepackage{subfigure} 

\usepackage{amssymb,amsmath}
\usepackage{mathtools}  
\mathtoolsset{showonlyrefs}

\DeclareMathOperator*{\supp}{supp}

\newcommand\bs[1]{\boldsymbol{#1}}
\newcommand\col[1]{\bar{\boldsymbol{#1}}}
\newcommand\tr{\mathrm{T}}
\newcommand\negvec[2]{\bs{#1}_{\neg #2}}
\newcommand\subvec[3]{\bs{#1}_{#2:#3}}
\newcommand\partialfrac[2]{\frac{\partial #1}{\partial #2}}
\newcommand\tensor[1]{\boldsymbol{\mathcal{#1}}}

\newcommand\anovakern[3]{\mathcal{A}^{#3}(\bs{#1}, \bs{#2})}
\newcommand\diagkern[3]{\mathcal{D}^{{#3}}(\bs{#1}, \bs{#2})}

\title{Higher-Order Factorization Machines}

\author{
  Mathieu Blondel, Akinori Fujino, Naonori Ueda \\
  NTT Communication Science Laboratories\\
  Japan \\
  \And
  Masakazu Ishihata \\
  Hokkaido University \\
  Japan
}

\begin{document}

\maketitle

\begin{abstract}
Factorization machines (FMs) are a supervised learning approach that can use
second-order feature combinations even when the data is very high-dimensional.
Unfortunately, despite increasing interest in FMs, there exists to date no
efficient training algorithm for higher-order FMs (HOFMs). In this paper, we
present the first generic yet efficient algorithms for training arbitrary-order
HOFMs. We also present new variants of HOFMs with shared parameters, which
greatly reduce model size and prediction times while maintaining similar
accuracy.  We demonstrate the proposed approaches on four different link
prediction tasks.
\end{abstract}

\section{Introduction}

Factorization machines (FMs) \cite{fm,libfm} are a supervised learning approach
that can use second-order feature combinations efficiently even when the data is
very high-dimensional. The key idea of FMs is to model the weights of feature
combinations using a \textit{low-rank} matrix. This has two main benefits.
First, FMs can achieve empirical accuracy on a par with polynomial regression or
kernel methods but with smaller and faster to evaluate models \cite{fm_icml}.
Second, FMs can infer the weights of feature combinations that were not observed
in the training set. This second property is crucial for instance in
recommender systems, a domain where FMs have become increasingly
popular \cite{libfm,fm_context}.  Without the low-rank property, FMs would fail
to generalize to unseen user-item interactions.  

Unfortunately, although higher-order FMs (HOFMs) were briefly mentioned
in the original work of \cite{fm,libfm}, there exists to date no efficient
algorithm for training arbitrary-order HOFMs. In fact, even just computing
predictions given the model parameters naively takes polynomial time in the
number of features.  For this reason, HOFMs have, to our knowledge, never been
applied to any problem.  In addition, HOFMs, as originally defined in
\cite{fm,libfm}, model each degree in the polynomial expansion with a different
matrix and therefore require the estimation of a large number of parameters.

In this paper, we propose the first efficient algorithms for training
arbitrary-order HOFMs. To do so, we rely on a link between FMs and the so-called
ANOVA kernel \cite{fm_icml}. We propose linear-time dynamic programming
algorithms for evaluating the ANOVA kernel and computing its gradient. Based on
these, we propose stochastic gradient and coordinate descent algorithms for
arbitrary-order HOFMs. To reduce the number of parameters, as well as prediction
times, we also introduce two new kernels derived from the ANOVA kernel, allowing
us to define new variants of HOFMs with shared parameters. We demonstrate the
proposed approaches on four different link prediction tasks.

\section{Factorization machines (FMs)}
\label{sec:fms}

{\bf Second-order FMs.} Factorization machines (FMs) \cite{fm,libfm} are an
increasingly popular method for efficiently using second-order feature
combinations in classification or regression tasks even when the data is very
high-dimensional.  Let $\bs{w} \in \mathbb{R}^d$ and $\bs{P} \in \mathbb{R}^{d
\times k}$, where $k \in \mathbb{N}$ is a rank hyper-parameter. We denote the
rows of $\bs{P}$ by $\col{p}_j$ and its columns by $\bs{p}_s$, for $j \in [d]$
and $s \in [k]$, where $[d] \coloneqq \{1,\dots,d\}$.  Then, FMs predict an
output $y \in \mathbb{R}$ from a vector $\bs{x} = [x_1, \dots, x_d]^\tr$
by
\begin{equation}
    \hat{y}_{\text{FM}}(\bs{x}) \coloneqq \langle \bs{w}, \bs{x} \rangle +
    \sum_{j'> j} \langle \col{p}_j, \col{p}_{j'} \rangle x_j x_{j'}.
\label{eq:predict_fm}
\end{equation}
An important characteristic of \eqref{eq:predict_fm} is that it considers only
combinations of \textit{distinct features} (i.e., the squared features
$x_1^2,\dots,x_d^2$ are ignored).  The main advantage of FMs compared to naive
polynomial regression is that the number of parameters to estimate is $O(dk)$
instead of $O(d^2)$. In addition, we can compute
predictions in $O(2dk)$ time\footnote{We include the constant factor for fair
later comparison with arbitrary-order HOFMs.} using 
\begin{equation}
\hat{y}_{\text{FM}}(\bs{x}) 
= \bs{w}^\tr \bs{x} + \frac{1}{2} \Big(\|\bs{P}^\tr
\bs{x}\|^2 - \sum_{s=1}^k \|\bs{p}_{s} \circ \bs{x}\|^2 \Big),
\label{eq:predict_fm_fast}
\end{equation}
where $\circ$ indicates element-wise product \cite{convex_fm}. 
Given a training set $\bs{X} = [\bs{x}_1, \dots,
\bs{x}_n] \in \mathbb{R}^{d \times n}$ and $\bs{y} = [y_1, \dots, y_n]^\tr \in
\mathbb{R}^n$, $\bs{w}$ and $\bs{P}$ can be learned by minimizing the following
non-convex objective
\begin{equation}
\frac{1}{n} \sum_{i=1}^n \ell\left(y_i, \hat{y}_{\text{FM}}(\bs{x}_i)\right) +
\frac{\beta_1}{2} \|\bs{w}\|^2 + \frac{\beta_2}{2} \|\bs{P}\|^2,
\label{eq:obj_fm}
\end{equation}
where $\ell$ is a convex loss function and $\beta_1 > 0, \beta_2 > 0$ are
hyper-parameters. The popular \texttt{libfm} library \cite{libfm} implements
efficient stochastic gradient and coordinate descent algorithms for obtaining a
stationary point of \eqref{eq:obj_fm}. Both algorithms have a runtime complexity 
of $O(2dkn)$ per epoch.

{\bf Higher-order FMs (HOFMs).}
Although no training algorithm was provided, FMs were extended to higher-order
feature combinations in the original work of \cite{fm,libfm}. Let
$\bs{P}^{(t)} \in \mathbb{R}^{d \times k_t}$, where $t \in \{2,\dots,m\}$ is the
order or degree of feature combinations considered, and $k_t \in \mathbb{N}$ is
a rank hyper-parameter.  Let $\col{p}_j^{(t)}$ be the $j^{\text{th}}$ row of
$\bs{P}^{(t)}$.  Then $m$-order HOFMs can be defined as
\begin{equation}
\hat{y}_{\text{HOFM}}(\bs{x}) \coloneqq \langle \bs{w}, \bs{x} \rangle +
\sum_{j'> j} \langle \col{p}_j^{(2)}, \col{p}_{j'}^{(2)} \rangle x_j x_{j'} 
+ \dots +
\sum_{j_m > \dots > j_1} \langle \col{p}_{j_1}^{(m)}, \dots, \col{p}_{j_m}^{(m)}
\rangle 
x_{j_1} x_{j_2} \dots x_{j_m}
\label{eq:predict_hofm}
\end{equation}
where we defined $\langle \col{p}_{j_1}^{(t)}, \dots, \col{p}_{j_t}^{(t)}
\rangle \coloneqq 
\text{sum}(\col{p}_{j_1}^{(t)} \circ \dots \circ \col{p}_{j_t}^{(t)})$ (sum of
element-wise products). The objective function of HOFMs can be expressed in a
similar way as for \eqref{eq:obj_fm}:
\begin{equation}
\frac{1}{n} \sum_{i=1}^n \ell\left(y_i, \hat{y}_{\text{HOFM}}(\bs{x}_i)\right) +
\frac{\beta_1}{2} \|\bs{w}\|^2 + \sum_{t=2}^m \frac{\beta_t}{2}
\|\bs{P}^{(t)}\|^2,
\label{eq:obj_hofm}
\end{equation}
where $\beta_1,\dots,\beta_m > 0$ are hyper-parameters. To avoid the
combinatorial explosion of hyper-parameter combinations to search, in our
experiments we will simply set $\beta_1 = \dots = \beta_m$ and $k_2 =
\dots = k_m$. While \eqref{eq:predict_hofm} looks quite
daunting, \cite{fm_icml} recently showed that FMs can be expressed from a
simpler kernel perspective. Let us define the ANOVA\footnote{The name comes from
the ANOVA decomposition of functions. \cite{wahba_book,vapnik_book}} kernel
\cite{vapnik_book} of degree $2 \le m \le d$ by
\begin{equation}
    \mathcal{A}^m(\bs{p}, \bs{x}) \coloneqq \sum_{j_m > \dots > j_1}
    \prod_{t=1}^m p_{j_t} x_{j_t}.
\label{eq:anova_kernel}
\end{equation}
For later convenience, we also define $\anovakern{p}{x}{0} \coloneqq 1$
and $\anovakern{p}{x}{1} \coloneqq \langle \bs{p}, \bs{x} \rangle$.
Then it is shown that 
\vspace{-0.1cm}
\begin{equation}
\hat{y}_{\text{HOFM}}(\bs{x}) = \langle \bs{w}, \bs{x} \rangle +
\sum_{s=1}^{k_2} \mathcal{A}^2\left(\bs{p}_s^{(2)}, \bs{x}\right) + \dots + 
\sum_{s=1}^{k_m} \mathcal{A}^m\left(\bs{p}_s^{(m)}, \bs{x}\right),
\label{eq:predict_hofm_anova}
\end{equation}
where $\bs{p}_s^{(t)}$ is the $s^{\text{th}}$ column of $\bs{P}^{(t)}$.  This
perspective shows that we can view FMs and HOFMs as a type of kernel machine
whose ``support vectors'' are learned directly from data.  Intuitively, the
ANOVA kernel can be thought as a kind of polynomial kernel that uses feature
combinations without replacement (i.e., of
\textit{distinct} features).  A key property of the ANOVA kernel
is multi-linearity
\cite{fm_icml}:
\begin{equation} 
\anovakern{p}{x}{m} = \mathcal{A}^m(\negvec{p}{j}, \negvec{x}{j}) + ~ p_j x_j ~
\mathcal{A}^{m-1}(\bs{p}_{\neg j}, \bs{x}_{\neg j}),
\label{eq:multi_linearity}
\end{equation}
where $\bs{p}_{\neg j}$ denotes the $(d-1)$-dimensional vector with $p_j$
removed and similarly for $\negvec{x}{j}$.
That is, everything else kept fixed, $\anovakern{p}{x}{m}$ is an affine function
of $p_j ~ \forall j \in [d]$.
Although no training algorithm was provided,
\cite{fm_icml} showed based on \eqref{eq:multi_linearity} that, although
non-convex, the objective
function of arbitrary-order HOFMs is convex in $\bs{w}$ and in
each row of $\bs{P}^{(2)},\dots,\bs{P}^{(m)}$, separately. 

{\bf Interpretability of HOFMs.} An advantage of FMs and HOFMs is their
interpretability. To see why this is
the case, notice that we can rewrite \eqref{eq:predict_hofm} as
\begin{equation}
\hat{y}_{\text{HOFM}}(\bs{x}) = \langle \bs{w}, \bs{x} \rangle +
\sum_{j'> j} \tensor{W}^{(2)}_{j,j'} x_j x_{j'} 
+ \dots +
\sum_{j_m > \dots > j_1} \tensor{W}^{(m)}_{j_1,\dots,j_m} x_{j_1} x_{j_2} \dots
x_{j_m},
\end{equation}
where we defined $\tensor{W}^{(t)} \coloneqq \sum_{s=1}^{k_t}
\underbrace{\bs{p}_s^{(t)} \otimes \dots \otimes \bs{p}_s^{(t)}}_{t \text{
times}}$. Intuitively, $\tensor{W}^{(t)} \in \mathbb{R}^{d^t}$ is a low-rank
$t$-way tensor which contains the weights of feature combinations of degree $t$.
For instance, when $t=3$,  $\tensor{W}^{(3)}_{i,j,k}$ is the weight of $x_i x_j
x_k$. 
Similarly to the ANOVA decomposition of functions, HOFMs consider only
combinations of distinct features (i.e., $x_{j_1} x_{j_2} \dots x_{j_m}$ for
$j_m > \dots > j_2 > j_1$).

{\bf This paper.}  Unfortunately, there exists to date no efficient algorithm
for training arbitrary-order HOFMs. Indeed, computing \eqref{eq:anova_kernel}
naively takes $O(d^m)$, i.e., polynomial time.  In the following, we present
linear-time algorithms. Moreover, HOFMs, as originally defined in
\cite{fm,libfm} require the estimation of $m-1$ matrices
$\bs{P}^{(2)},\dots,\bs{P}^{(m)}$. Thus, HOFMs can produce large models when $m$
is large. To address this issue, we propose new variants of HOFMs with shared
parameters.

\section{Linear-time stochastic gradient algorithms for HOFMs}
\label{sec:anova_sgd}

The kernel view presented in Section \ref{sec:fms} allows us to focus on the
ANOVA kernel as the main ``computational unit'' for training HOFMs.  In this
section, we develop dynamic programming (DP) algorithms for evaluating the ANOVA
kernel and computing its gradient in only $O(dm)$ time. 

{\bf Evaluation.} The main observation (see also \cite[Section
9.2]{kernel_book}) is that we can use
\eqref{eq:multi_linearity} to recursively remove features until computing the
kernel becomes trivial. Let us denote a subvector of $\bs{p}$ by
$\subvec{p}{1}{j} \in \mathbb{R}^j$ and similarly for $\bs{x}$. Let
us introduce the shorthand $a_{j,t} \coloneqq \mathcal{A}^t(\subvec{p}{1}{j},
\subvec{x}{1}{j})$.  Then, from \eqref{eq:multi_linearity}, 
\begin{equation} 
a_{j,t} = a_{j-1,t} + ~ p_j x_j ~ a_{j-1,t-1} \quad \forall d \ge j \ge t \ge 1.
\label{eq:recursion1}
\end{equation}
For convenience, we also define $a_{j,0} = 1 ~ \forall j \ge 0$ since
$\anovakern{p}{x}{0}=1$ and $a_{j,t} = 0 ~ \forall j < t$ since there does not
exist any $t$-combination of features in a $j < t$ dimensional vector.

\begin{wraptable}{r}{5.8cm}
\vspace{-10pt}
\caption{\footnotesize Example of DP table}
\scriptsize
\begin{tabular}{c|ccccc}
& $j=0$ & $j=1$ & $j=2$ & $\dots$ & $j=d$ \\
\hline
$t=0$ & 1 & $1$ & $1$ & $1$ & $1$ \\
$t=1$ & 0 & $a_{1,1}$ & $a_{2,1}$ & $\dots$ & $a_{d,1}$\\
$t=2$ & 0 & $0$ & $a_{2,2}$ & $\dots$ & $a_{d,2}$\\
$\vdots$ & $\vdots$ & $\vdots$ & $\vdots$ & $\ddots$ & $\vdots$ \\
$t=m$ & 0 & $0$ & $0$ & $\dots$ & $a_{d,m}$ \\
\end{tabular}
\end{wraptable}

The quantity we want to compute is $\anovakern{p}{x}{m} = a_{d,m}$. Instead of
naively using recursion \eqref{eq:recursion1}, which would lead to many redundant
computations, we use a bottom-up approach and organize computations in
a DP table. We start from the top-left corner to initialize the
recursion and go through the table to arrive at the solution in the bottom-right
corner.  The procedure, summarized in Algorithm
\ref{algo:anova_eval}, takes $O(dm)$ time and memory.

{\bf Gradients.} For computing the gradient of $\anovakern{p}{x}{m}$ w.r.t.
$\bs{p}$, we use reverse-mode differentiation \cite{autodiff} (a.k.a.
backpropagation in a neural network context), since it allows us to compute the
entire gradient in a single pass. We supplement each variable $a_{j,t}$ in the
DP table by a so-called adjoint $\tilde{a}_{j,t} \coloneqq
\partialfrac{a_{d,m}}{a_{j,t}}$, which represents the sensitivity of $a_{d,m} =
\anovakern{p}{x}{m}$ w.r.t.  $a_{j,t}$. From recursion \eqref{eq:recursion1},
except for edge cases, $a_{j,t}$ influences $a_{j+1,t+1}$ and $a_{j+1,t}$.
Using the chain rule, we then obtain
\begin{equation}
\tilde{a}_{j,t} = 
\partialfrac{a_{d,m}}{a_{j+1,t}} \partialfrac{a_{j+1,t}}{a_{j,t}}
+
\partialfrac{a_{d,m}}{a_{j+1,t+1}} \partialfrac{a_{j+1,t+1}}{a_{j,t}}
= 
\tilde{a}_{j+1,t}
+
p_{j+1} x_{j+1} ~ \tilde{a}_{j+1,t+1}
\quad \forall d-1 \ge j \ge t \ge 1.
\label{eq:backprop_recursion}
\end{equation}
Similarly, we introduce the adjoint $\tilde{p}_j \coloneqq
\partialfrac{a_{d,m}}{p_j} ~ \forall j \in [d]$. Since $p_j$ influences $a_{j,t}
~ \forall t \in [m]$, we have
\begin{equation}
    \tilde{p}_j = \sum_{t=1}^m \partialfrac{a_{d,m}}{a_{j,t}}
    \partialfrac{a_{j,t}}{p_j} = \sum_{t=1}^m \tilde{a}_{j,t} ~ a_{j-1, t-1} ~
    x_j.
\end{equation}
We can run recursion \eqref{eq:backprop_recursion} in reverse order of the DP
table starting from $\tilde{a}_{d,m} = \partialfrac{a_{d,m}}{a_{d,m}} = 1$.
Using this approach, we can compute the entire gradient $\nabla
\anovakern{p}{x}{m} = [\tilde{p}_1, \dots, \tilde{p}_d]^\tr$ w.r.t. $\bs{p}$ in
$O(dm)$ time and memory.  The procedure is summarized in Algorithm
\ref{algo:anova_grad}.

\begin{figure*}[t]
\begin{minipage}[t]{0.48\linewidth} 
\begin{algorithm}[H]
    \caption{\footnotesize Evaluating $\anovakern{p}{x}{m}$ in $O(dm)$}
\begin{algorithmic}
\footnotesize
\STATE {\bfseries Input:} $\bs{p} \in \mathbb{R}^d$, $\bs{x} \in \mathbb{R}^d$
\STATE $a_{j,t} \leftarrow 0 ~ \forall t \in [m], j \in
    [d] \cup \{0\}$
    \STATE $a_{j,0} \leftarrow 1 ~ \forall j \in [d] \cup \{0\}$
\vspace{1em}
\FOR{$t \coloneqq 1, \dots, m$}
    \FOR{$j \coloneqq t, \dots, d$}
    \STATE $a_{j,t} \leftarrow a_{j-1,t} + p_j x_j
        a_{j-1,t-1}$
    \ENDFOR
\ENDFOR
\vspace{1em}
\STATE {\bfseries Output:} $\anovakern{p}{x}{m} = a_{d,m}$
\end{algorithmic}
\label{algo:anova_eval}
\end{algorithm}
\end{minipage}
\begin{minipage}[t]{0.48\linewidth} 
\begin{algorithm}[H]
\caption{\footnotesize Computing $\nabla \anovakern{p}{x}{m}$ in $O(dm)$}
\begin{algorithmic}
\footnotesize
\STATE {\bfseries Input:} $\bs{p} \in \mathbb{R}^d$, $\bs{x} \in \mathbb{R}^d$,
$\{a_{j,t}\}_{j,t=0}^{d,m}$
\STATE $\tilde{a}_{j,t} \leftarrow 0 ~ \forall t \in [m+1], j \in [d]$
\STATE $\tilde{a}_{d,m} \leftarrow 1$
\vspace{0.28em}
\FOR{$t \coloneqq m, \dots, 1$}
    \FOR{$j \coloneqq d-1, \dots, t$}
    \STATE $\tilde{a}_{j,t} \leftarrow 
    \tilde{a}_{j+1,t} + \tilde{a}_{j+1,t+1} p_{j+1} x_{j+1}$
    \ENDFOR
\ENDFOR
\vspace{0.28em}
\STATE $\tilde{p}_j \coloneqq \sum_{t=1}^m \tilde{a}_{j,t} a_{j-1,t-1} x_j ~
\forall j \in [d]$
\STATE {\bfseries Output:} $\nabla \anovakern{p}{x}{m} = [\tilde{p}_1, \dots,
\tilde{p}_d]^\tr$
\end{algorithmic}
\label{algo:anova_grad}
\end{algorithm}
\end{minipage}
\end{figure*}

{\bf Stochastic gradient (SG) algorithms.} Based on Algorithm
\ref{algo:anova_eval} and \ref{algo:anova_grad}, we can easily learn
arbitrary-order HOFMs using any gradient-based optimization algorithm. Here we
focus our discussion on SG algorithms.  If we alternatingly minimize
\eqref{eq:obj_hofm} w.r.t $\bs{P}^{(2)}, \dots, \bs{P}^{(m)}$, then the
sub-problem associated with degree $m$ is of the form
\begin{equation}
    F(\bs{P}) \coloneqq \frac{1}{n} \sum_{i=1}^n \ell\left(y_i, \sum_{s=1}^k
    \mathcal{A}^m(\bs{p}_s, \bs{x}_i) + o_i \right) + \frac{\beta}{2}
    \|\bs{P}\|^2,
\label{eq:obj_anova}
\end{equation}
where $o_1,\dots,o_n \in \mathbb{R}$ are fixed offsets which account for the
contribution of degrees other than $m$ to the predictions.  The sub-problem is
convex in each row of $\bs{P}$ \cite{fm_icml}. A SG update for
\eqref{eq:obj_anova} w.r.t. $\bs{p}_s$ for some instance $\bs{x}_i$ can be
computed by $\bs{p}_s \leftarrow \bs{p}_s - \eta \ell'(y_i,\hat{y}_i) \nabla
\mathcal{A}^m(\bs{p}_s, \bs{x}_i) - \eta \beta \bs{p}_s$, where $\eta$ is a
learning rate and where we defined $\hat{y}_i \coloneqq \sum_{s=1}^k
\mathcal{A}^m(\bs{p}_s, \bs{x}_i) + o_i$. Because evaluating
$\anovakern{p}{x}{m}$ and computing its gradient both take $O(dm)$, the cost
per epoch, i.e.,  of visiting all instances, is $O(mdkn)$. When $m=2$, this
is the same cost as the SG algorithm implemented in
\texttt{libfm}. 

{\bf Sparse data.} We conclude this section with a few useful remarks on sparse
data.  Let us denote the support of a vector $\bs{\bs{x}} = [x_1, \dots,
x_d]^\tr$ by $\supp(\bs{x}) \coloneqq \{j \in [d] \colon x_j \neq 0\}$ and let
us define $\bs{x}_S \coloneqq [x_j \colon j \in S]^\tr$. It is easy to see from
\eqref{eq:multi_linearity} that the gradient and $\bs{x}$ have the same support,
i.e., $\supp(\nabla \anovakern{p}{x}{m}) = \supp(\bs{x})$. Another useful remark
is that $\anovakern{p}{x}{m} = \mathcal{A}^m(\bs{p}_{\supp(\bs{x})},
\bs{x}_{\supp(\bs{x})})$, provided that $m \le n_z(\bs{x})$, where $n_z(\bs{x})$
is the number of non-zero elements in $\bs{x}$.  Hence, when the data is sparse,
we only need to iterate over non-zero features in Algorithm
\ref{algo:anova_eval} and \ref{algo:anova_grad}.  Consequently, their time and
memory cost is only $O(n_z(\bs{x}) m)$ and thus the cost per epoch of SG
algorithms is $O(mk n_z(\bs{X}))$. 

\section{Coordinate descent algorithm for arbitrary-order HOFMs}
\label{sec:anova_cd}

We now describe a coordinate descent (CD) solver for arbitrary-order HOFMs. CD
is a good choice for learning HOFMs because their objective function is
coordinate-wise convex, thanks to the multi-linearity of the ANOVA kernel
\cite{fm_icml}.  Our algorithm can be seen as a generalization to higher orders
of the CD algorithms proposed in \cite{libfm, fm_icml}.

{\bf An alternative recursion.} Efficient CD implementations typically require
maintaining statistics for each training instance, such as the predictions at
the current iteration. When a coordinate is updated, the statistics then need to
be synchronized.  Unfortunately, the recursion we used in the previous section
is not suitable for a CD algorithm because it would require to store and
synchronize the DP table for each training instance upon coordinate-wise
updates. We therefore turn to an alternative recursion:
\begin{equation}
\anovakern{p}{x}{m} = \frac{1}{m} \sum_{t=1}^m (-1)^{t+1}
\anovakern{p}{x}{m-t} \diagkern{p}{x}{t},
\label{eq:recursion2}
\end{equation}
where we defined $\diagkern{p}{x}{t} \coloneqq \sum_{j=1}^d (p_j x_j)^t$.  Note
that the recursion was already known in the context of traditional kernel
methods (c.f., \cite[Section 11.8]{vapnik_book}) but its application to HOFMs is
novel.  Since we know that $\anovakern{p}{x}{0}=1$ and $\anovakern{p}{x}{1} =
\langle \bs{p}, \bs{x} \rangle$, we can use \eqref{eq:recursion2} to compute
$\anovakern{p}{x}{2}$, then $\anovakern{p}{x}{3}$, and so on.  The overall
evaluation cost for arbitrary $m \in \mathbb{N}$ is $O(md + m^2)$.

{\bf Coordinate-wise derivatives.} We can apply reverse-mode differentiation to
recursion \eqref{eq:recursion2} in order to compute the entire gradient
(c.f., Appendix \ref{appendix:alt_dp_gradient}).  However, in CD, since we only
need the derivative of one variable at a time, we can simply use forward-mode
differentiation:
\begin{equation}
\partialfrac{\anovakern{p}{x}{m}}{p_j} =
\frac{1}{m} \sum_{t=1}^m (-1)^{t+1} \left[
\partialfrac{\anovakern{p}{x}{m-t}}{p_j} \diagkern{p}{x}{t} +
\anovakern{p}{x}{m-t} \partialfrac{\diagkern{p}{x}{t}}{p_j} \right],
\label{eq:coordinate_wise_derivative}
\end{equation}
where $\partialfrac{\diagkern{p}{x}{t}}{p_j} = t p_j^{t-1} x_j^t$.  The
advantage of \eqref{eq:coordinate_wise_derivative} is that we only need to cache
$\mathcal{D}^t(\bs{p}, \bs{x})$ for $t \in [m]$. Hence the memory complexity per
sample is only $O(m)$ instead of $O(dm)$ for \eqref{eq:recursion1}.

{\bf Use in a CD algorithm.} Similarly to \cite{fm_icml}, we assume that the
loss function $\ell$ is $\mu$-smooth and update the elements $p_{j,s}$ of
$\bs{P}$ in cyclic order by $p_{j,s} \leftarrow p_{j,s} - \eta_{j,s}^{-1}
    \partialfrac{F(\bs{P})}{p_{j,s}}$, where we defined
\begin{align}
\eta_{j,s} \coloneqq \frac{\mu}{n} \sum_{i=1}^n
\left(\partialfrac{\mathcal{A}^m(\bs{p}_s, \bs{x}_i)}{p_{j,s}}\right)^2 + \beta
\quad \text{and} \quad
\partialfrac{F(\bs{P})}{p_{j,s}} = \frac{1}{n} \sum_{i=1}^n \ell'(y_i,
\hat{y}_i) \partialfrac{\mathcal{A}^m(\bs{p}_s, \bs{x}_i)}{p_{j,s}} + \beta
p_{j,s}.
\end{align}
The update guarantees that the objective value is monotonically non-increasing
and is the exact coordinate-wise minimizer when $\ell$ is the squared loss.
Overall, the total cost per epoch, i.e., updating all coordinates once, is
$O(\tau(m)kn_z(\bs{X}))$, where $\tau(m)$ is the time it takes to compute
\eqref{eq:coordinate_wise_derivative}.  Assuming $\mathcal{D}^t(\bs{p}_s,
\bs{x}_i)$ have been previously cached, for $t \in [m]$,
computing \eqref{eq:coordinate_wise_derivative} takes $\tau(m) = m (m + 1) / 2 -
1$ operations.  For fixed $m$, if we unroll the two loops needed to compute
\eqref{eq:coordinate_wise_derivative}, modern compilers can often further reduce
the number of operations needed.  Nevertheless, this quadratic dependency  on
$m$ means that our CD algorithm is best for small $m$, typically $m \le 4$.

\vspace{-0.1cm}
\section{HOFMs with shared parameters}
\label{sec:anova_shared}

HOFMs, as originally defined in \cite{fm,libfm}, model each degree with
\textit{separate} matrices $\bs{P}^{(2)}, \dots, \bs{P}^{(m)}$. Assuming that we
use the same rank $k$ for all matrices, the total model size of $m$-order HOFMs
is therefore $O(kdm)$.  Moreover, even when using our $O(dm)$ DP algorithm, the
cost of computing predictions is $O(k(2d + \dots + md)) = O(kdm^2)$. Hence,
HOFMs tend to produce large, expensive-to-evaluate models. To reduce model size
and prediction times, we introduce two new kernels
which allow us to \textit{share} parameters between each degree: the
\textbf{inhomogeneous ANOVA kernel} and the \textbf{all-subsets kernel}. Because
both kernels are derived from the ANOVA kernel, they share the same appealing
properties: multi-linearity, sparse gradients and sparse-data friendliness.

\subsection{Inhomogeneous ANOVA kernel}
\label{sec:inhomo_anova}

It is well-known that a sum of kernels is equivalent to concatenating their
associated feature maps \cite[Section 3.4]{kernel_book}.  Let $\bs{\theta} =
[\theta_1, \dots, \theta_m]^\tr$. To combine different degrees, a natural kernel
is therefore
\begin{equation}
\mathcal{A}^{1 \rightarrow m}(\bs{p}, \bs{x}; \bs{\theta}) \coloneqq 
\sum_{t=1}^m \theta_t \anovakern{p}{x}{t}.
\label{eq:inhomogeneous_anova}
\end{equation}
The kernel uses all feature combinations of degrees $1$ up to $m$.  We call it
\textit{inhomogeneous} ANOVA kernel, since it is an inhomogeneous polynomial of
$\bs{x}$. In contrast, $\anovakern{p}{x}{m}$ is homogeneous. The main difference
between \eqref{eq:inhomogeneous_anova} and \eqref{eq:predict_hofm_anova} is that
all ANOVA kernels in the sum share the same parameters. However, to increase
modeling power, we allow each kernel to have different weights $\theta_1, \dots,
\theta_m$.  

{\bf Evaluation.} Due to the recursive nature of Algorithm
\ref{algo:anova_eval}, when computing $\anovakern{p}{x}{m}$, we also get
$\anovakern{p}{x}{1}, \dots, \anovakern{p}{x}{m-1}$ for free. Indeed,
lower-degree kernels are available in the last column of the DP table, i.e.,
$\anovakern{p}{x}{t} = a_{d,t} ~ \forall t \in [m]$. Hence, the cost of
evaluating \eqref{eq:inhomogeneous_anova} is $O(dm)$ time.
The total cost for computing $\hat{y} = \sum_{s=1}^k \mathcal{A}^{1 \rightarrow
m}(\bs{p}_s, \bs{x}; \bs{\theta})$ is $O(kdm)$ instead of $O(kdm^2)$ for 
$\hat{y}_{\text{HOFM}}(\bs{x})$.

{\bf Learning.} While it is certainly possible to learn $\bs{P}$ and
$\bs{\theta}$ by directly minimizing some objective function, here we propose an
easier solution, which works well in practice.  Our key observation is that we
can easily turn $\mathcal{A}^m$ into $\mathcal{A}^{1 \rightarrow m}$ by adding
dummy values to feature vectors. Let us denote the concatenation of $\bs{p}$
with a scalar $\gamma$ by $[\gamma, \bs{p}]$ and similarly for $\bs{x}$. 
From \eqref{eq:multi_linearity}, we easily obtain
\begin{equation}
\mathcal{A}^m([\gamma_1, \bs{p}], [1, \bs{x}]) =
\anovakern{p}{x}{m} + \gamma_1 \anovakern{p}{x}{m-1}.
\end{equation}
Similarly, if we apply \eqref{eq:multi_linearity} twice, we obtain:
\begin{equation}
\mathcal{A}^m([\gamma_1, \gamma_2, \bs{p}], [1, 1, \bs{x}]) =
\anovakern{p}{x}{m} + (\gamma_1 + \gamma_2) \anovakern{p}{x}{m-1} + \gamma_1
\gamma_2 \anovakern{p}{x}{m-2}.
\end{equation}
Applying the above to $m=2$ and $m=3$, we obtain
\begin{small}
\begin{equation}
\mathcal{A}^2([\gamma_1, \bs{p}], [1, \bs{x}]) = 
\mathcal{A}^{1 \rightarrow 2}(\bs{p}, \bs{x}; [\gamma_1, 1])
\quad \text{and} \quad
\mathcal{A}^3([\gamma_1, \gamma_2, \bs{p}], [1, 1, \bs{x}]) =
\mathcal{A}^{1 \rightarrow 3}(\bs{p}, \bs{x}; [\gamma_1 \gamma_2, \gamma_1 +
\gamma_2, 1]).
\end{equation}
\end{small}
More generally, by adding $m-1$ dummy features to $\bs{p}$ and $\bs{x}$, we can
convert $\mathcal{A}^m$ to $\mathcal{A}^{1 \rightarrow m}$. Because $\bs{p}$ is
learned, this means that we can automatically learn $\gamma_1, \dots,
\gamma_{m-1}$. These weights can then be converted to
$\theta_1,\dots,\theta_m$ by ``unrolling'' recursion \eqref{eq:multi_linearity}.
Although simple, we show in our experiments that this approach works favorably
compared to directly learning $\bs{P}$ and $\bs{\theta}$. The main advantage of
this approach is that we can use the same software unmodified (we simply need to
minimize \eqref{eq:obj_anova} with the augmented data). Moreover, the
cost of computing the entire gradient by Algorithm \ref{algo:anova_grad} using
the augmented data is just $O(dm+m^2)$ compared to $O(dm^2)$ for 
HOFMs with separate parameters.

\subsection{All-subsets kernel}
\label{sec:all_subsets}

We now consider a closely related kernel called all-subsets kernel
\cite[Definition 9.5]{kernel_book}:
\begin{equation}
\mathcal{S}(\bs{p}, \bs{x}) \coloneqq \prod_{j=1}^d (1 + p_j x_j).
\label{eq:all_subset_kernel}
\end{equation}
The main difference with the traditional use of this kernel is that we learn
$\bs{p}$.
Interestingly, it can be shown that $\mathcal{S}(\bs{p}, \bs{x}) = 1 +
\mathcal{A}^{1 \rightarrow d}(\bs{p}, \bs{x}; \bs{1}) = 1 + \mathcal{A}^{1
\rightarrow n_z(\bs{x})}(\bs{p}, \bs{x}; \bs{1})$, where $n_z(\bs{x})$ is the
number of non-zero features in $\bs{x}$.  Hence, the kernel uses
\textit{all} combinations of distinct features up to order
$n_z(\bs{x})$ \textit{with uniform weights}.  Even if $d$ is very large, the
kernel can be a good choice if each training instance contains only a few
non-zero elements. To learn the parameters, we simply substitute $\mathcal{A}^m$
with $\mathcal{S}$ in \eqref{eq:obj_anova}.  In SG or CD algorithms, all it
entails is to substitute $\nabla \anovakern{p}{x}{m}$ with $\nabla
\mathcal{S}(\bs{p}, \bs{x})$.  For computing $\nabla
\mathcal{S}(\bs{p}, \bs{x})$, it is
easy to verify that $\mathcal{S}(\bs{p}, \bs{x}) = \mathcal{S}(\bs{p}_{\neg j},
\bs{x}_{\neg j}) (1 + p_j x_j) ~ \forall j \in [d]$ and therefore we have
\begin{equation}
\nabla \mathcal{S}(\bs{p}, \bs{x}) =
\bigg[x_1 ~ \mathcal{S}(\negvec{p}{1}, \negvec{x}{1}), \dots,
x_d ~ \mathcal{S}(\negvec{p}{d}, \negvec{x}{d}) \bigg]^\tr
=
\bigg[\frac{x_1 ~ \mathcal{S}(\bs{p}, \bs{x})}{1 + p_1 x_1}, \dots,
\frac{x_d ~ \mathcal{S}(\bs{p}, \bs{x})}{1 + p_d x_d}\bigg]^\tr.
\end{equation}
Therefore, the main advantage of the all-subsets kernel is that we can evaluate
it and compute its gradient in just $O(d)$ time.
The total cost for computing $\hat{y} = \sum_{s=1}^k \mathcal{S}
(\bs{p}_s, \bs{x})$ is only $O(kd)$.

\vspace{-0.1cm}
\section{Experimental results}

\subsection{Application to link prediction}
\label{sec:link_prediction}

{\bf Problem setting.} We now demonstrate a novel application of HOFMs to
predict the presence or absence of links between nodes in a graph.  Formally, we
assume two sets of possibly disjoint nodes of size $n_A$ and $n_B$,
respectively.  We assume features for the two sets of nodes, represented by
matrices $\bs{A} \in \mathbb{R}^{d_A \times n_A}$ and $\bs{B} \in
\mathbb{R}^{d_B \times n_B}$. For instance, $\bs{A}$ can represent user
features and $\bs{B}$ movie features.  We denote the columns of $\bs{A}$ and
$\bs{B}$ by $\bs{a}_i$ and $\bs{b}_j$, respectively.  We are given a matrix
$\bs{Y} \in \{0,1\}^{n_A \times n_B}$, whose elements indicate presence
(positive sample) or absence (negative sample) of link between two nodes
$\bs{a}_i$ and $\bs{b}_j$. We denote the number of positive samples by $n_+$.
Using this data, our goal is to predict new associations.
Datasets used in our experiments are summarized in Table \ref{table:datasets}.
Note that for the NIPS and Enzyme datasets, $\bs{A}=\bs{B}$.

\begin{table}[t]
\caption{Datasets used in our experiments. For a detailed 
description, c.f. Appendix
\protect\ref{appendix:dataset_description}.}
\begin{center}
\begin{small}
    \begin{tabular}{c|c|c|c|c|c|c|c}
Dataset & $n_+$ & Columns of $\bs{A}$ & $n_A$ & $d_A$ & Columns of $\bs{B}$ & $n_B$ & $d_B$ \\
\hline
NIPS \cite{nips_data} & 4,140 & Authors & 2,037 & 13,649 &  &  & \\
Enzyme \cite{yamanishi_2005} & 2,994 & Enzymes & 668 & 325 & & & \\
GD \cite{imc} & 3,954 & Diseases & 3,209 & 3,209 & Genes & 12,331 & 25,275 \\
Movielens 100K \cite{movielens} & 21,201 & Users & 943 & 49 & Movies & 1,682 &
29 \\
\end{tabular}
\end{small}
\end{center}
\label{table:datasets}
\end{table}

{\bf Conversion to a supervised problem.} We need to convert the above
information to a format FMs and HOFMs can handle. To predict an element
$y_{i,j}$ of $\bs{Y}$, we simply form $\bs{x}_{i,j}$ to be the concatenation
of $\bs{a}_i$ and $\bs{b}_j$ and feed this to a HOFM in order to compute a prediction
$\hat{y}_{i,j}$. Because HOFMs use feature combinations in 
$\bs{x}_{i,j}$, they can learn the weights of feature combinations between
$\bs{a}_i$ and $\bs{b}_j$. At training time, we need both positive and negative
samples.  Let us denote the set of positive and negative
samples by $\Omega$. Then our training set is composed of $(\bs{x}_{i,j},
y_{i,j})$ pairs, where $(i,j) \in \Omega$. 

{\bf Models compared.}

\begin{itemize}[topsep=0pt,itemsep=1ex,partopsep=0ex,parsep=0.3ex,leftmargin=2ex]

\item \textit{HOFM}: $\hat{y}_{i,j} = \hat{y}_{\text{HOFM}}(\bs{x}_{i,j})$
as defined in \eqref{eq:predict_hofm} and as originally proposed in
\cite{fm,libfm}. We minimize \eqref{eq:obj_hofm} by alternating minimization of
\eqref{eq:obj_anova} for each degree. 

\item \textit{HOFM-shared}: $\hat{y}_{i,j} = \sum_{s=1}^k \mathcal{A}^{1
\rightarrow m}(\bs{p}_s, \bs{x}_{i,j}; \bs{\theta})$. We learn $\bs{P}$ and
$\bs{\theta}$ using the simple augmented data approach described in Section
\ref{sec:inhomo_anova} (\textit{HOFM-shared-augmented}).
Inspired by SimpleMKL \cite{simple_mkl}, we also report results when
learning $\bs{P}$ and $\bs{\theta}$ directly by minimizing
$\frac{1}{|\Omega|} \sum_{(i,j) \in \Omega} \ell(y_{i,j},
\hat{y}_{i,j})
+ \frac{\beta}{2} \|\bs{P}\|^2$ subject to $\bs{\theta} \ge 0$ and
$\langle \bs{\theta}, \bs{1} \rangle = 1$ (\textit{HOFM-shared-simplex}).

\item \textit{All-subsets}: $\hat{y}_{i,j} = \sum_{s=1}^k \mathcal{S}(\bs{p}_s,
\bs{x}_{i,j})$.  As explained in Section \ref{sec:all_subsets}, this model is
equivalent to the HOFM-shared model with $m=n_z(\bs{x}_{i,j})$ and
$\bs{\theta}=\bs{1}$.

\item \textit{Polynomial network}: $\hat{y}_{i,j} = \sum_{s=1}^k (\gamma_s +
\langle \bs{p}_s, \bs{x}_{i,j} \rangle)^m$. This model can be thought as
factorization machine variant that uses a polynomial kernel instead of the ANOVA
kernel (c.f., \cite{livni,fm_icml,tensor_machines}).

\item \textit{Low-rank bilinear regression}: $\hat{y}_{i,j} = \bs{a}_i \bs{U}
\bs{V}^\tr \bs{b}_j$, where $\bs{U} \in \mathbb{R}^{d_A \times k}$ and $\bs{V}
\in \mathbb{R}^{d_B \times k}$. Such model was shown to work well for
link prediction in \cite{link_prediction} and \cite{imc}.
We learn $\bs{U}$ and $\bs{V}$ by minimizing $\frac{1}{|\Omega|} \sum_{(i,j) \in
\Omega} \ell(y_{i,j}, \hat{y}_{i,j}) + \frac{\beta}{2} (\|\bs{U}\|^2 +
\|\bs{V}\|^2)$.

\end{itemize}

{\bf Experimental setup and evaluation.} In this experiment, for all models
above, we use CD rather than SG to avoid the tuning of a learning rate
hyper-parameter. We set $\ell$ to be the squared loss.  Although we omitted it
from our notation for clarity, we also fit a bias term for all models.  We
evaluated the compared models using the area under the ROC curve (AUC), which is
the probability that the model correctly ranks a positive sample higher than a
negative sample.  We split the $n_+$ positive samples into $50\%$ for training and
$50\%$ for testing. We sample the same number of negative samples as positive
samples for training and use the rest for testing.  We chose $\beta$ from
$10^{-6}, 10^{-5}, \dots, 10^{6}$ by cross-validation and following
\cite{link_prediction} we empirically set $k=30$.  Throughout our experiments,
we initialized the elements of $\bs{P}$ randomly by $\mathcal{N}(0, 0.01)$. 

\begin{table}[t]
\caption{Comparison of area under the ROC curve (AUC) as measured on the test
sets.}
\begin{center}
\begin{footnotesize}
    \begin{tabular}{c|c|c|c|c}
    & NIPS & Enzyme & GD & Movielens 100K \\
\hline
HOFM ($m=2$) & 0.856 & 0.880 & 0.717 & 0.778 \\
HOFM ($m=3$) & {\bf 0.875} & {\bf 0.888} & 0.717 & 0.786 \\
HOFM ($m=4$) & 0.874 & 0.887 & 0.717 & 0.786 \\
HOFM ($m=5$) & 0.874 & 0.887 & 0.717 & 0.786 \\
\hline
HOFM-shared-augmented ($m=2$) & 0.858 & 0.876 & 0.704 & 0.778 \\
HOFM-shared-augmented ($m=3$) & 0.874 & 0.887 & 0.704 & {\bf 0.787} \\
HOFM-shared-augmented ($m=4$) & 0.836 & 0.824 & 0.663 & 0.779 \\
HOFM-shared-augmented ($m=5$) & 0.824 & 0.795 & 0.600 & 0.621 \\
\hline
HOFM-shared-simplex ($m=2$) & 0.716 & 0.865 & {\bf 0.721} & 0.701 \\
HOFM-shared-simplex ($m=3$) & 0.777 & 0.870 & {\bf 0.721} & 0.709 \\
HOFM-shared-simplex ($m=4$) & 0.758 & 0.870 & {\bf 0.721} & 0.709 \\
HOFM-shared-simplex ($m=5$) & 0.722 & 0.869 & {\bf 0.721} & 0.709 \\
\hline
All-subsets & 0.730 & 0.840 & {\bf 0.721} & 0.714 \\
\hline
Polynomial network ($m=2$) & 0.725 & 0.879 & {\bf 0.721} & 0.761\\
Polynomial network ($m=3$) & 0.789 & 0.853 & 0.719 & 0.696 \\
Polynomial network ($m=4$) & 0.782 & 0.873 & 0.717 & 0.708 \\
Polynomial network ($m=5$) & 0.543 & 0.524 & 0.648 & 0.501 \\
\hline
Low-rank bilinear regression & 0.855 & 0.694 & 0.611 & 0.718 \\
\hline
\end{tabular}
\end{footnotesize}
\end{center}
\label{table:link_prediction}
\end{table}

{\bf Results} are indicated in Table \ref{table:link_prediction}.  Overall the
two best models were HOFM and HOFM-shared-augmented, which achieved the best
scores on 3 out of 4 datasets. The two models outperformed low-rank bilinear
regression on 3 out 4 datasets, showing the benefit of using higher-order
feature combinations. HOFM-shared-augmented achieved similar accuracy to HOFM,
despite using a smaller model. Surprisingly, HOFM-shared-simplex did not improve
over HOFM-shared-augmented except on the GD dataset. We conclude that our
augmented data approach is convenient yet works well in practice.  All-subsets
and polynomial networks performed worse than HOFM and HOFM-shared-augmented,
except on the GD dataset where they were the best. Finally, we observe that HOFM
were quite robust to increasing $m$, which is likely a benefit of modeling each
degree with a separate matrix.

\vspace{-0.3cm}
\subsection{Solver comparison}

\begin{figure}[t]
\center
\subfigure[Convergence when $m=2$]{
    \includegraphics[scale=0.19]{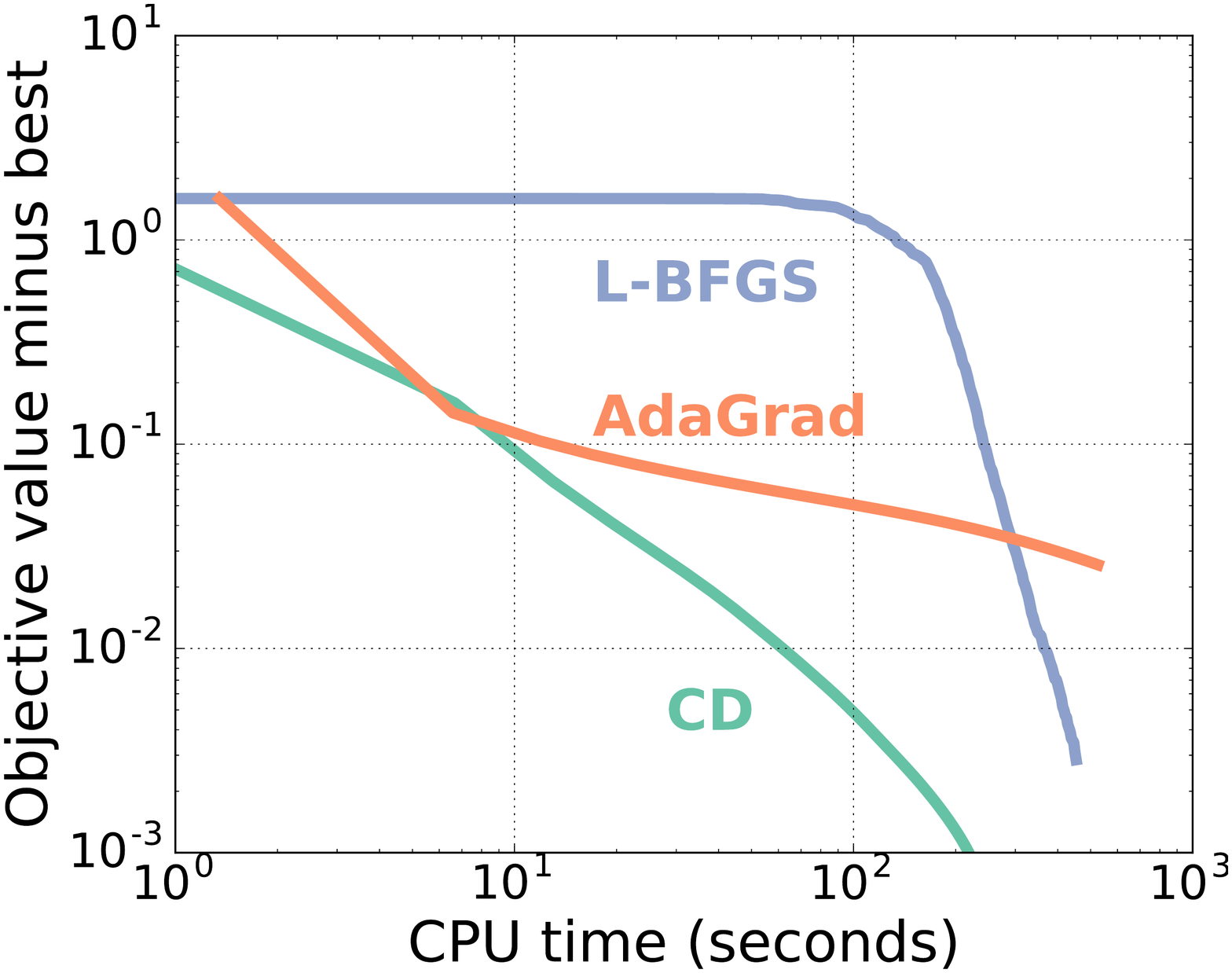}
}
\subfigure[Convergence when $m=3$]{
    \includegraphics[scale=0.19]{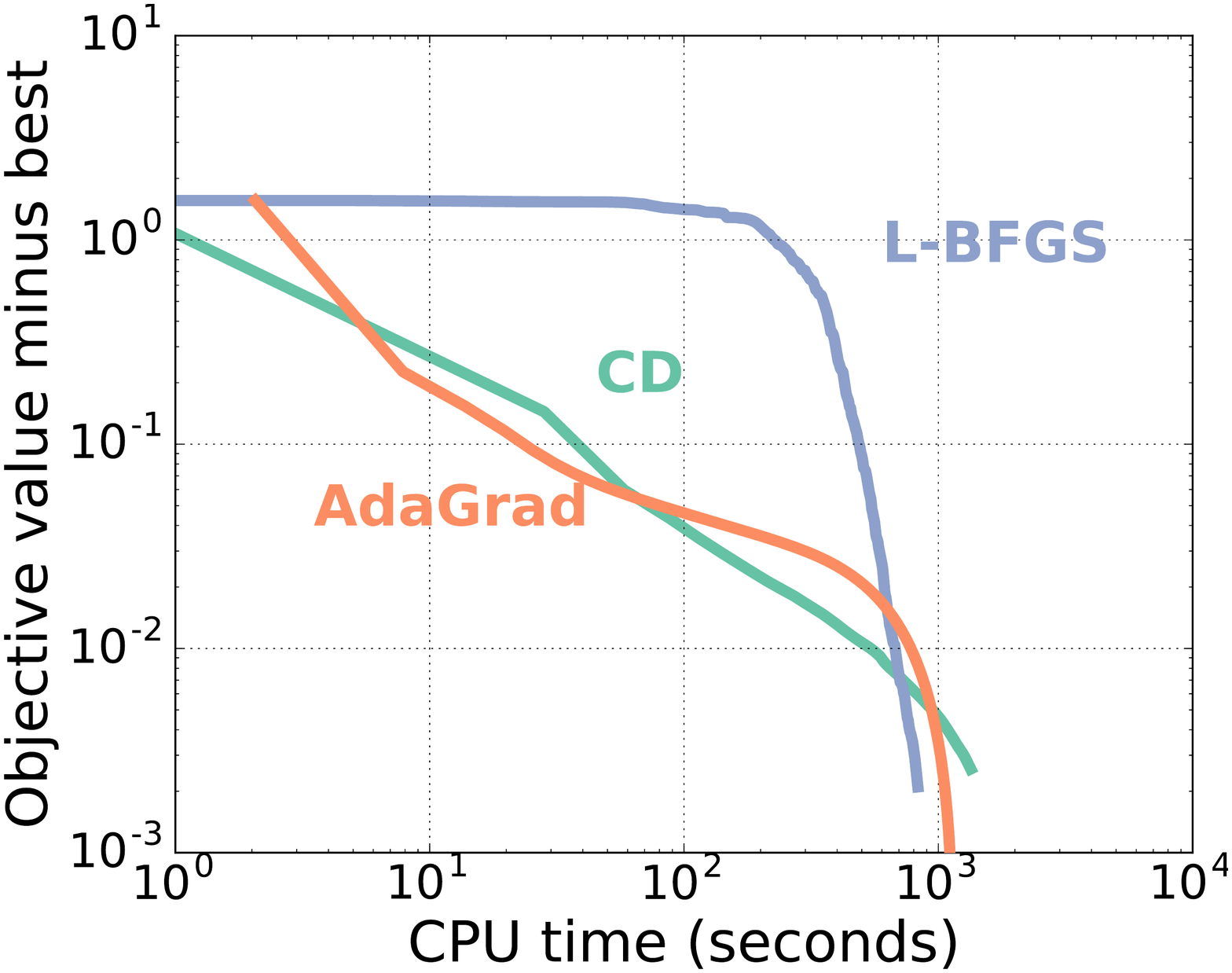}
}
\subfigure[Convergence when $m=4$]{
    \includegraphics[scale=0.19]{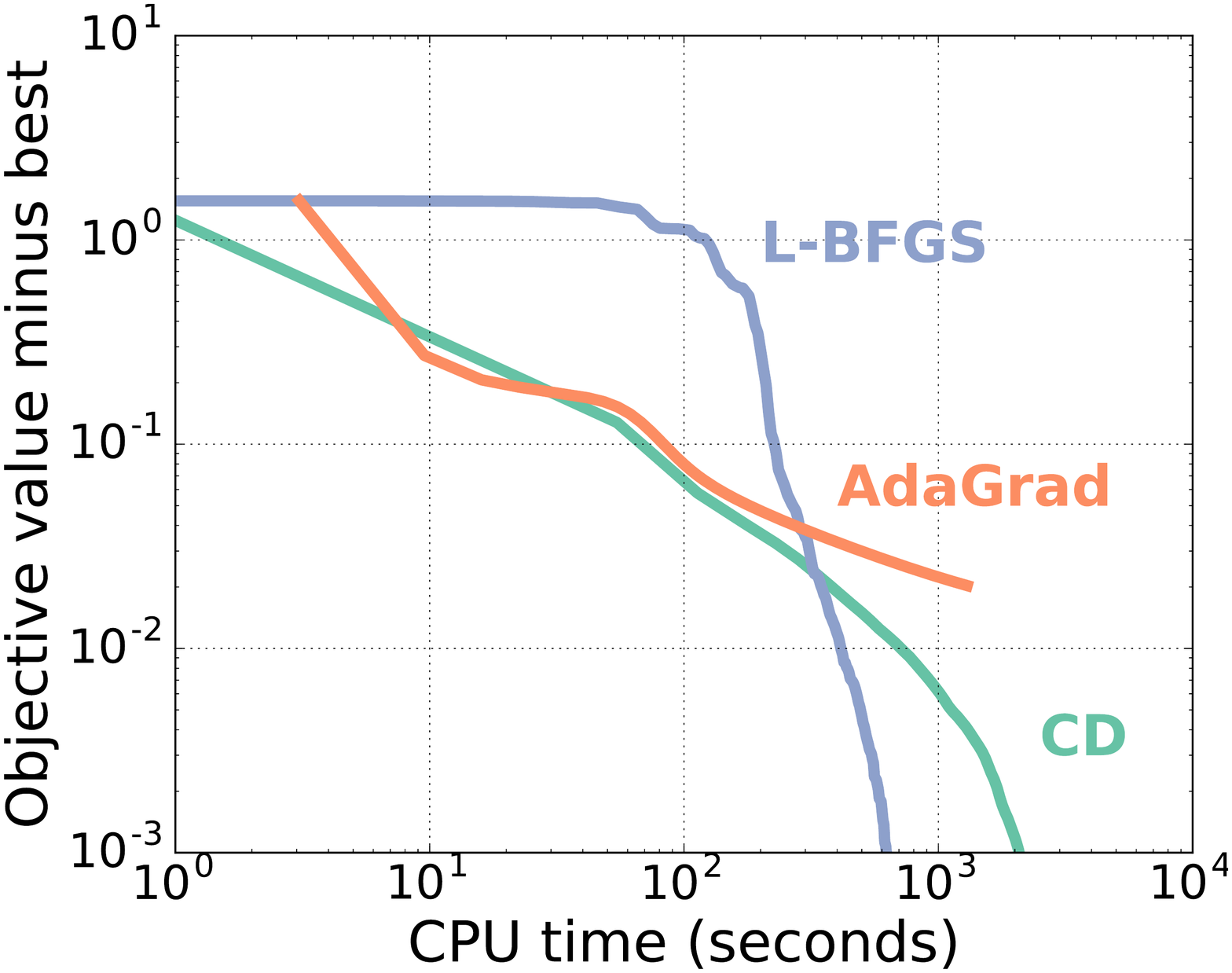}
}
\subfigure[Scalability w.r.t. degree $m$]{
    \includegraphics[scale=0.19]{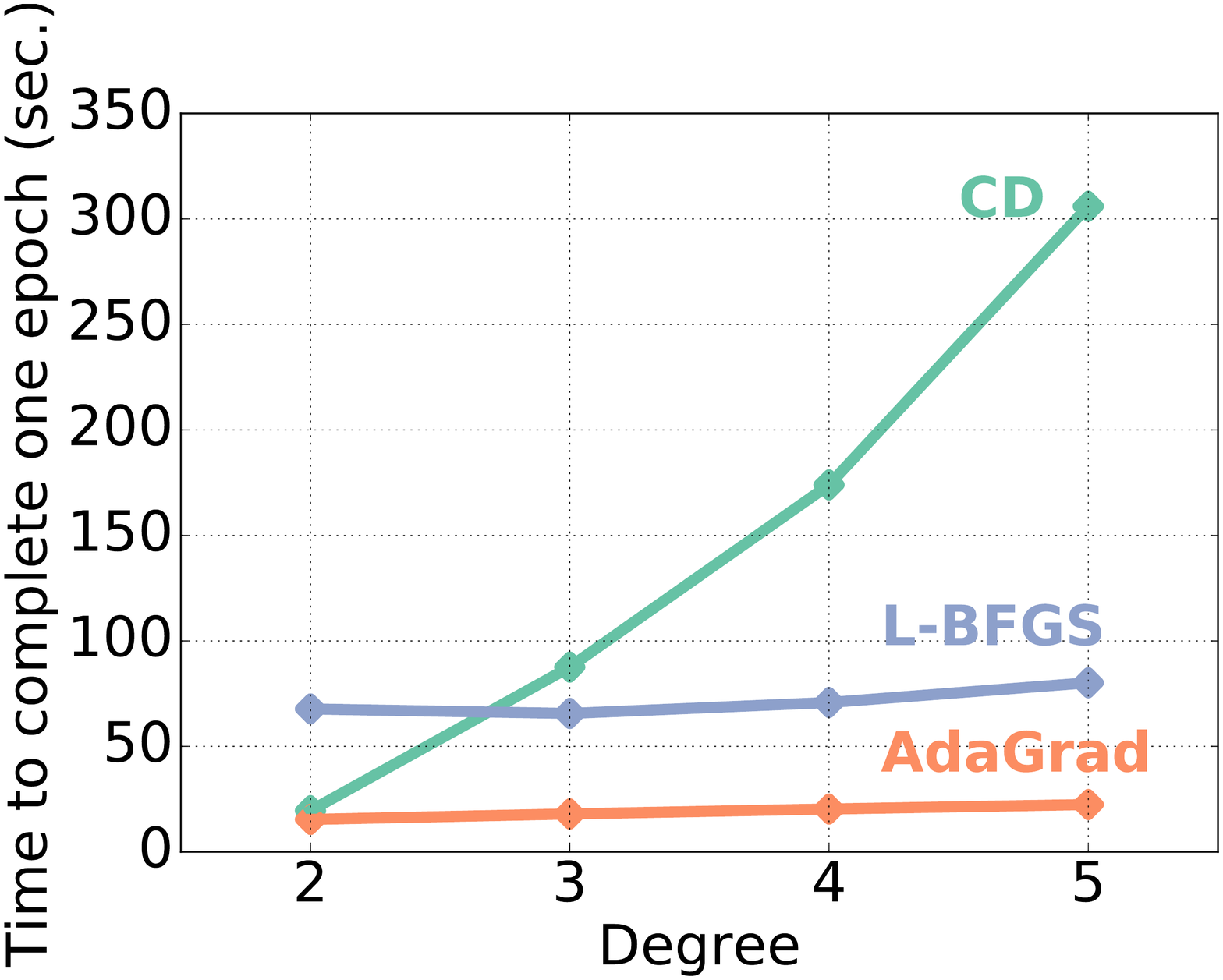}
}
\caption{Solver comparison for minimizing \protect\eqref{eq:obj_anova} when
varying the degree $m$ on the NIPS dataset with $\beta=0.1$ and $k=30$. Results
on other datasets are in Appendix
\protect\ref{appendix:additional_experiments}.}
\label{figure:solver_cmp}
\end{figure}

We compared AdaGrad \cite{adagrad}, L-BFGS and coordinate descent (CD) for
minimizing \eqref{eq:obj_anova} when varying the degree $m$ on the NIPS dataset
with $\beta=0.1$ and $k=30$.  We constructed the data in the same way as
explained in the previous section and added $m-1$ dummy features, resulting in
$n=8,280$ sparse samples of dimension $d=27,298 + m - 1$. For AdaGrad and
L-BFGS, we computed the (stochastic) gradients using Algorithm
\ref{algo:anova_grad}.  All solvers used the same initialization.

{\bf Results} are indicated in Figure \ref{figure:solver_cmp}.  We see that our
CD algorithm performs very well when $m \le 3$ but starts to deteriorate when $m
\ge 4$, in which case L-BFGS becomes advantageous.  As shown in Figure
\ref{figure:solver_cmp} d), the cost per epoch of AdaGrad and L-BFGS scales
linearly with $m$, a benefit of our DP algorithm for computing the gradient.
However, to
our surprise, we found that AdaGrad is quite sensitive to the learning rate
$\eta$.  AdaGrad diverged for $\eta \in \{1, 0.1, 0.01\}$ and the largest value
to work well was $\eta=0.001$. This explains why  AdaGrad did not outperform CD
despite the lower cost per epoch. 
In the future, it would be
useful to create a CD algorithm with a better dependency on $m$.

\vspace{-0.3cm}
\section{Conclusion and future directions}
\vspace{-0.3cm}

In this paper, we presented the first training algorithms for HOFMs and
introduced new HOFM variants with shared parameters.  A popular way to deal with
a large number of negative samples is to use an objective function that directly
maximize AUC \cite{link_prediction,rendle_bpr}.  This is especially easy to do
with SG algorithms because we can sample pairs of positive and negative samples
from the dataset upon each SG update. We therefore expect the algorithms
developed in Section \ref{sec:anova_sgd} to be especially useful in this
setting. Recently, \cite{difacto} proposed a distributed SG algorithm for
training second-order FMs. It should be straightforward to extend this algorithm
to HOFMs based on our contributions in Section \ref{sec:anova_sgd}. 
Finally, it should be possible to integrate Algorithm \ref{algo:anova_eval} and
\ref{algo:anova_grad} into a deep learning framework such as TensorFlow
\cite{tensorflow}, in order to easily compose
ANOVA kernels with other layers (e.g., convolutional).



\clearpage
\bibliography{paper_nips}
\bibliographystyle{abbrv}

\clearpage
\appendix

\begin{center}
    {\Huge \bf Supplementary material}
\end{center}

\section{Dataset descriptions}
\label{appendix:dataset_description}

\begin{itemize}[topsep=0pt,itemsep=1ex,partopsep=0ex,parsep=1ex,leftmargin=2ex]

\item \textbf{NIPS}: co-author graph of authors at the first twelve editions of
NIPS, obtained from \cite{nips_data}.  For this dataset, as well as the Enzyme
dataset below, we have $\bs{A}=\bs{B}$.  The co-author graph comprises $n_A =
n_B = 2,037$ authors represented by bag-of-words vectors of dimension $d_A = d_B
= 13,649$ (words used by authors in their publications). The number of positive
samples is $n_+ = 4,140$.

\item \textbf{Enzyme}: metabolic network obtained from \cite{yamanishi_2005}.
The network comprises $n_A = n_B = 668$ enzymes represented by three sets of
features: a $157$-dimensional vector of phylogenetic information, a
$145$-dimensional vector of gene expression information and a
$23$-dimensional vector of gene location information.  We concatenate the
three sets of information to form feature vectors of dimension $d_A = d_B =
325$. Original enzyme similarity scores are between $0$ and $1$. We binarize
the scores using $0.95$ as threshold.  The resulting number of positive
samples is $n_+ = 2,994$.

\item \textbf{GD}: human gene-disease association data obtained from
\cite{imc}. The bipartite graph is comprised of $n_A = 3,209$ diseases and $n_B
= 12,331$ genes. We represent each disease using a vector of $d_A = 3,209$
dimensions, whose elements are similarity scores obtained from MimMiner.  The
study \cite{imc} also used bag-of-words vectors describing each disease but we
found these to not help improve performance both for baselines and proposed
methods. We represent each gene using a vector of $d_B = 25,275$ features,
which are the concatenation of $12,331$ similarity scores obtained from
HumanNet and $12,944$ gene-phenotype associations from 8 other species. See
\cite{imc} for a detailed description of the features. The number of positive
samples is $n_+ = 3,954$.

\item \textbf{Movielens 100K}: recommender system data obtained from
\cite{movielens}.  The bipartite graph is comprised of $n_A = 943$ users and
$n_B = 1,682$ movies.  For users, we convert age, gender, occupation and living
area (first digit of zipcode) to a binary vector using a one-hot encoding. For
movies, we use the release year and genres. The resulting vectors are of
dimension $d_A=49$ and $d_B=29$, respectively.  Original ratings are between $1$
and $5$. We binarize the ratings using $5$ as threshold, resulting in $n_+ =
21,201$ positive samples.

\end{itemize}

\section{Additional experiments}
\label{appendix:additional_experiments}

\subsection{Solver comparison}

\begin{figure}[p]
\center
\subfigure[$m=2$]{
    \includegraphics[scale=0.14]{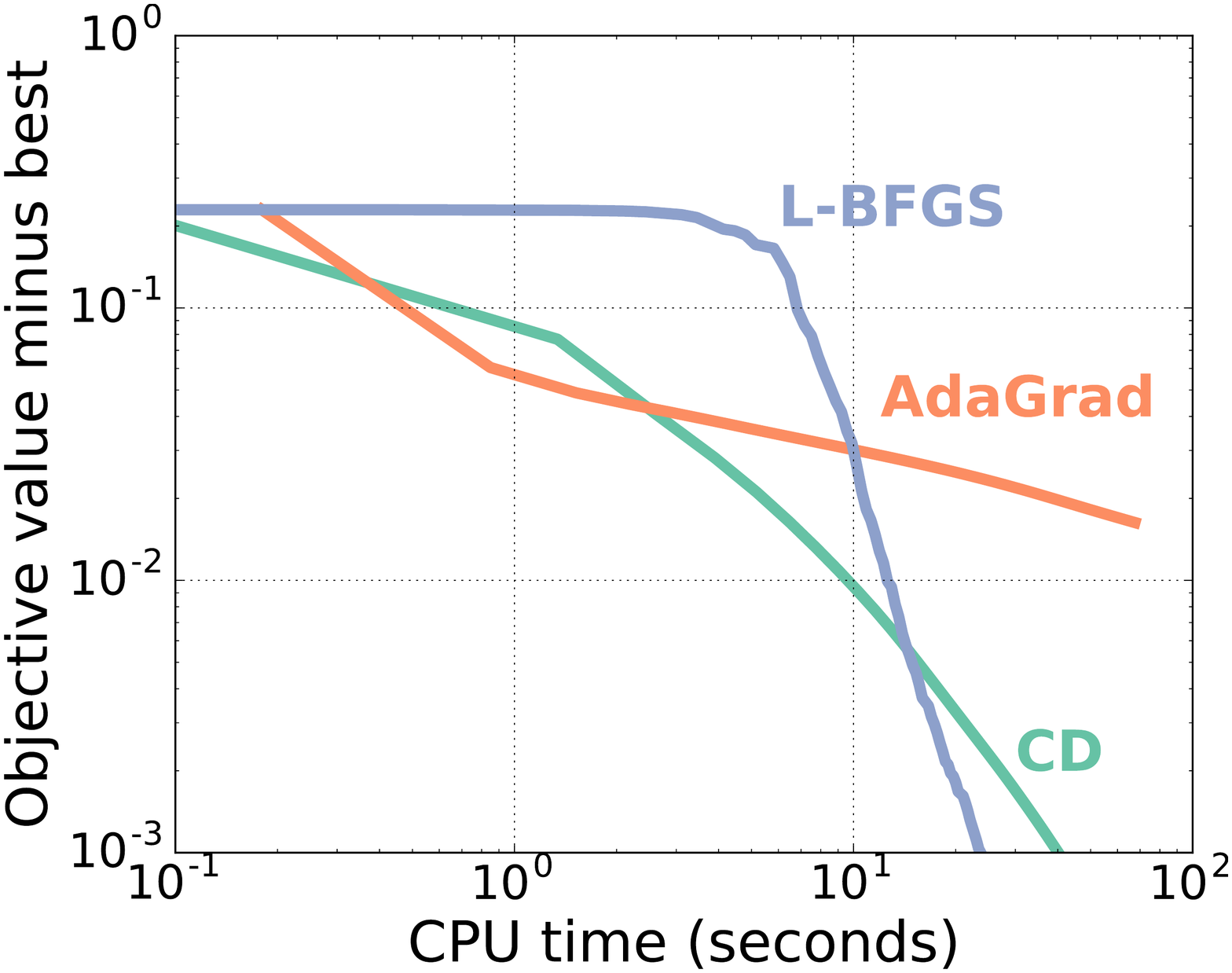}
}
\subfigure[$m=3$]{
    \includegraphics[scale=0.14]{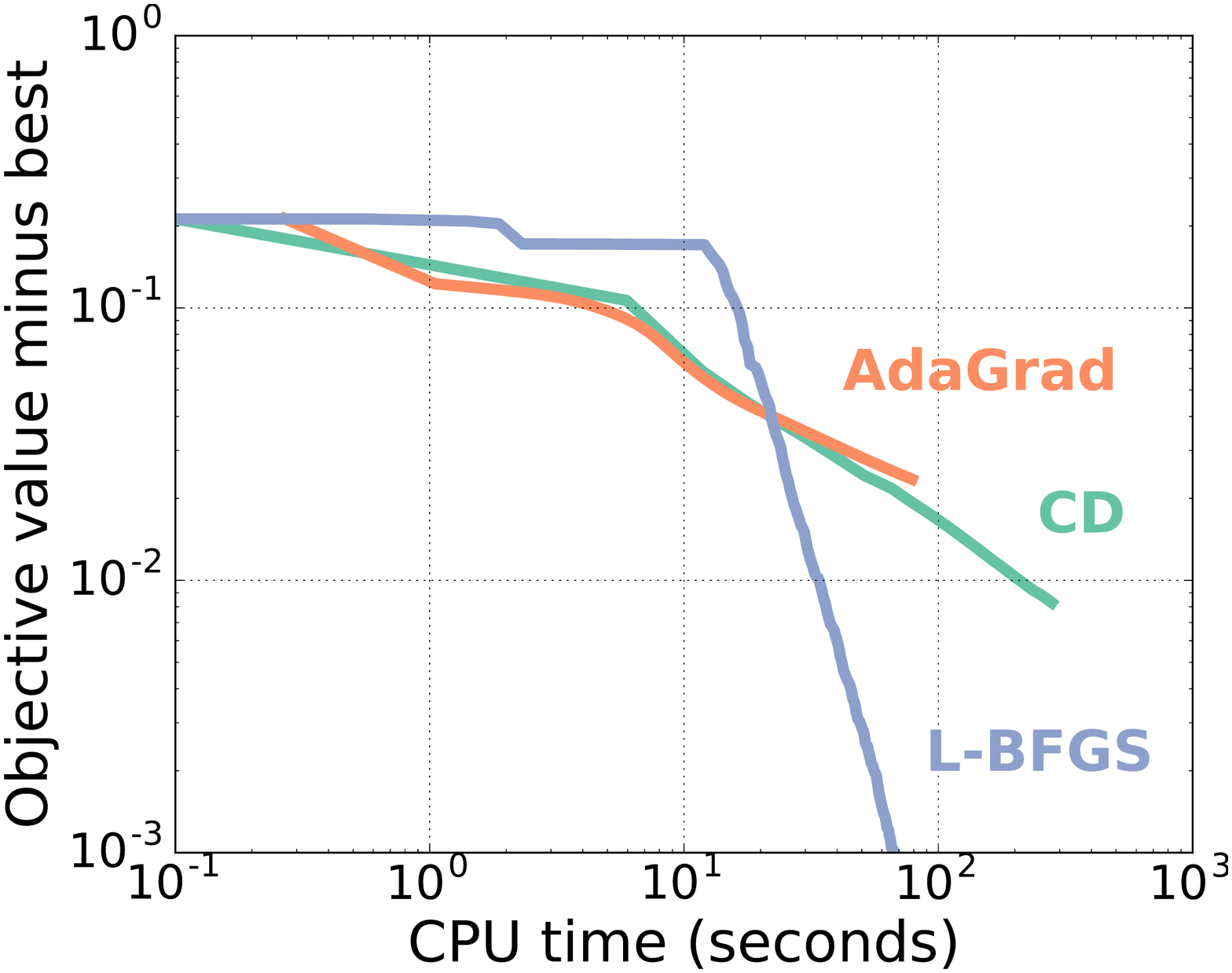}
}
\subfigure[$m=4$]{
    \includegraphics[scale=0.14]{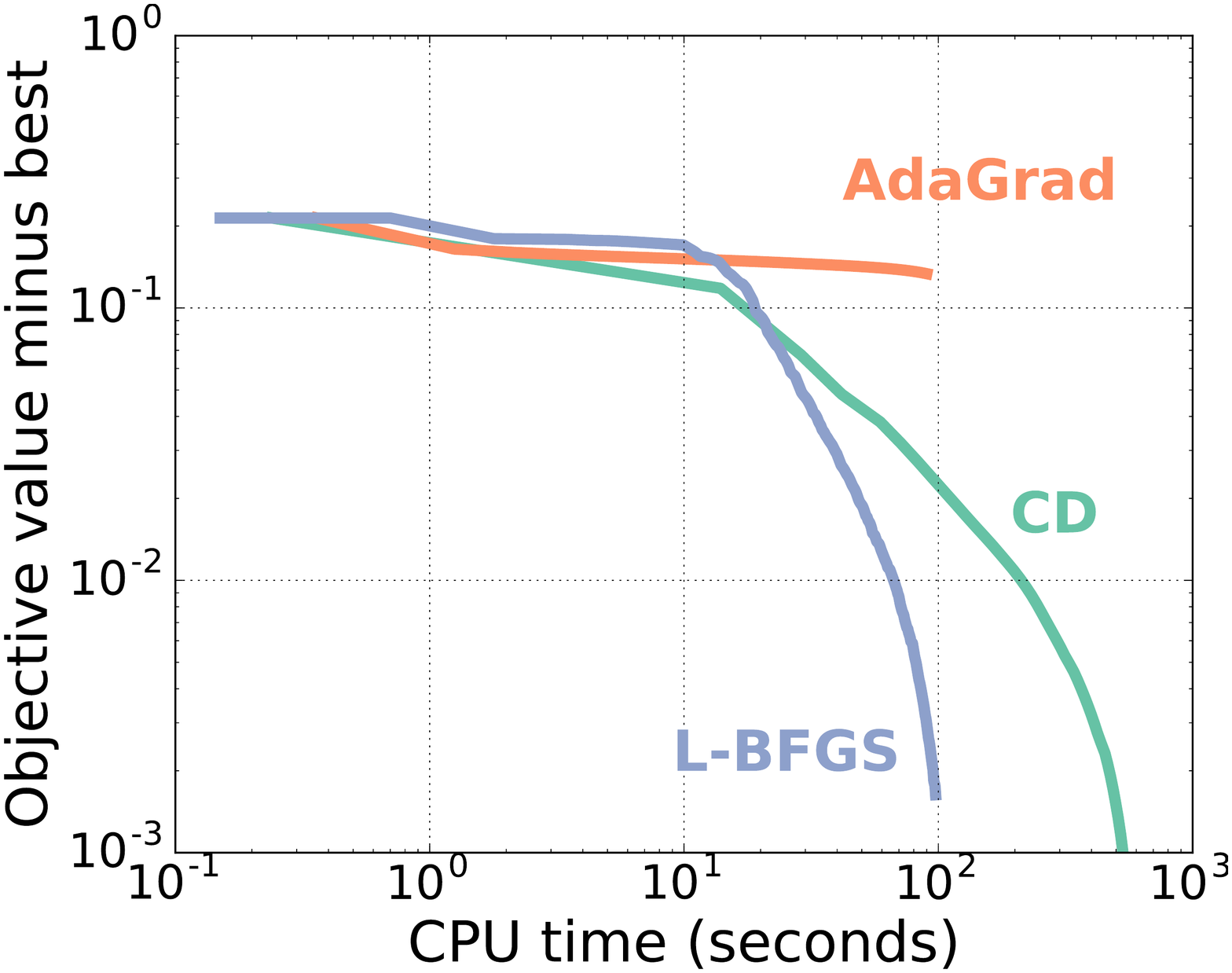}
}
\caption{Solver comparison for minimizing \protect\eqref{eq:obj_anova} on the
Enzyme dataset. We set $\beta$ to the values with best test-set performance,
which were $\beta=0.1$, $\beta=0.1$ and $\beta=0.01$, respectively. We set
$k=30$.} \label{figure:solver_cmp2}
\end{figure}

\begin{figure}[p]
\center
\subfigure[$m=2$]{
    \includegraphics[scale=0.14]{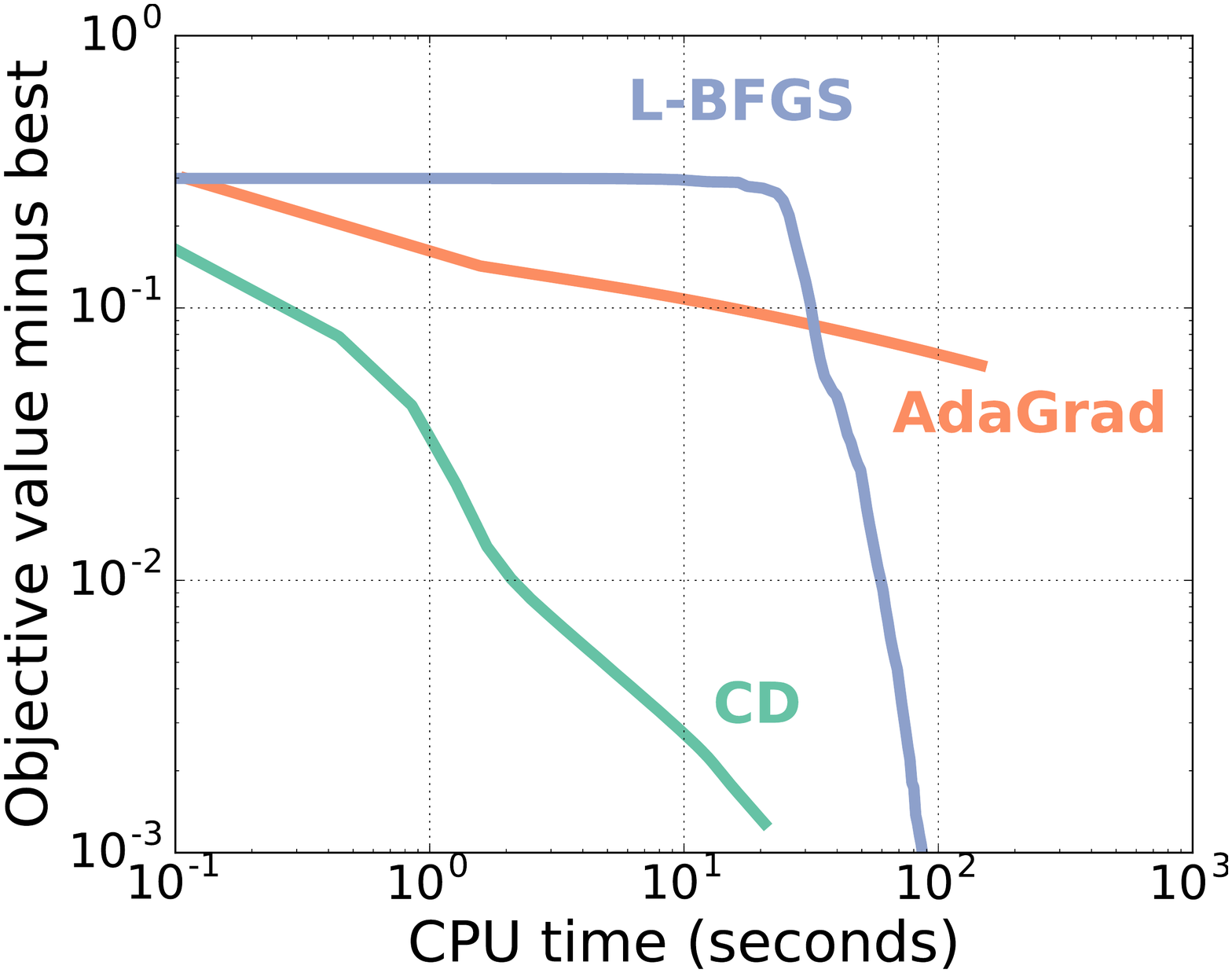}
}
\subfigure[$m=3$]{
    \includegraphics[scale=0.14]{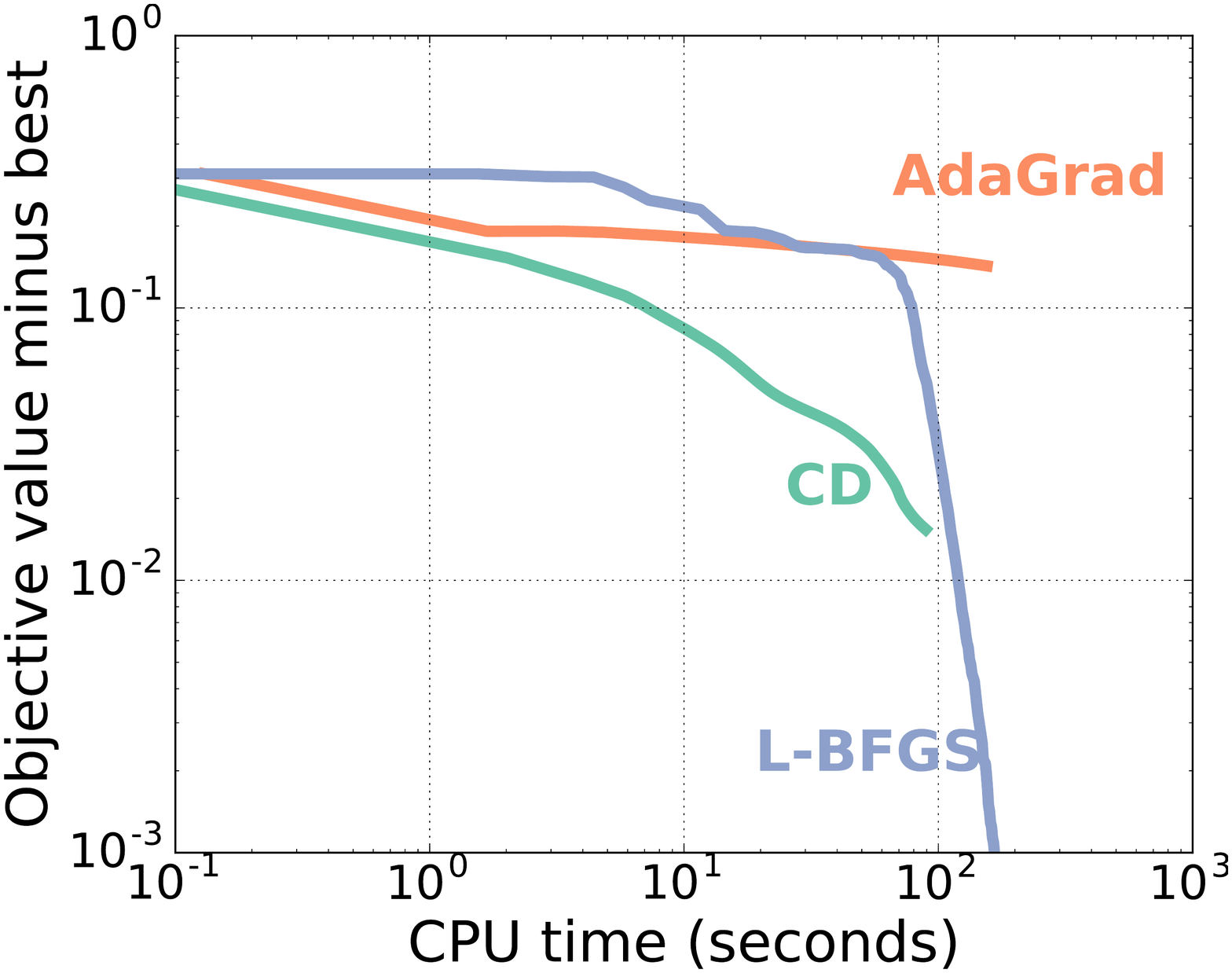}
}
\subfigure[$m=4$]{
    \includegraphics[scale=0.14]{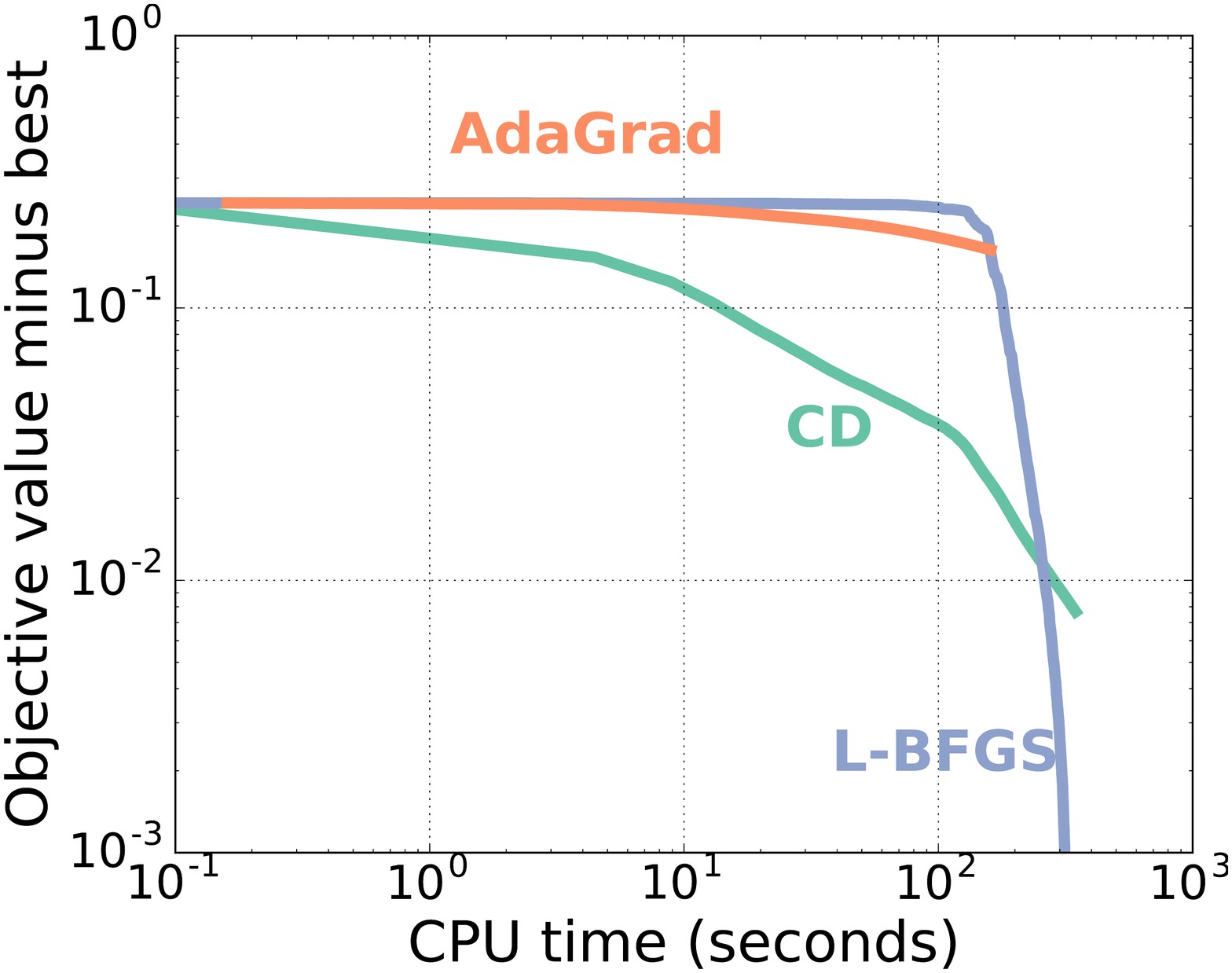}
}
\caption{Solver comparison for minimizing \protect\eqref{eq:obj_anova} on the GD
dataset. We set $\beta$ to the values with best test-set performance, which were
$\beta=0.01$, $\beta=0.01$ and $\beta=0.0001$, respectively. We set $k=30$.}
\label{figure:solver_cmp3}
\end{figure}

\begin{figure}[p]
\center
\subfigure[$m=2$]{
    \includegraphics[scale=0.14]{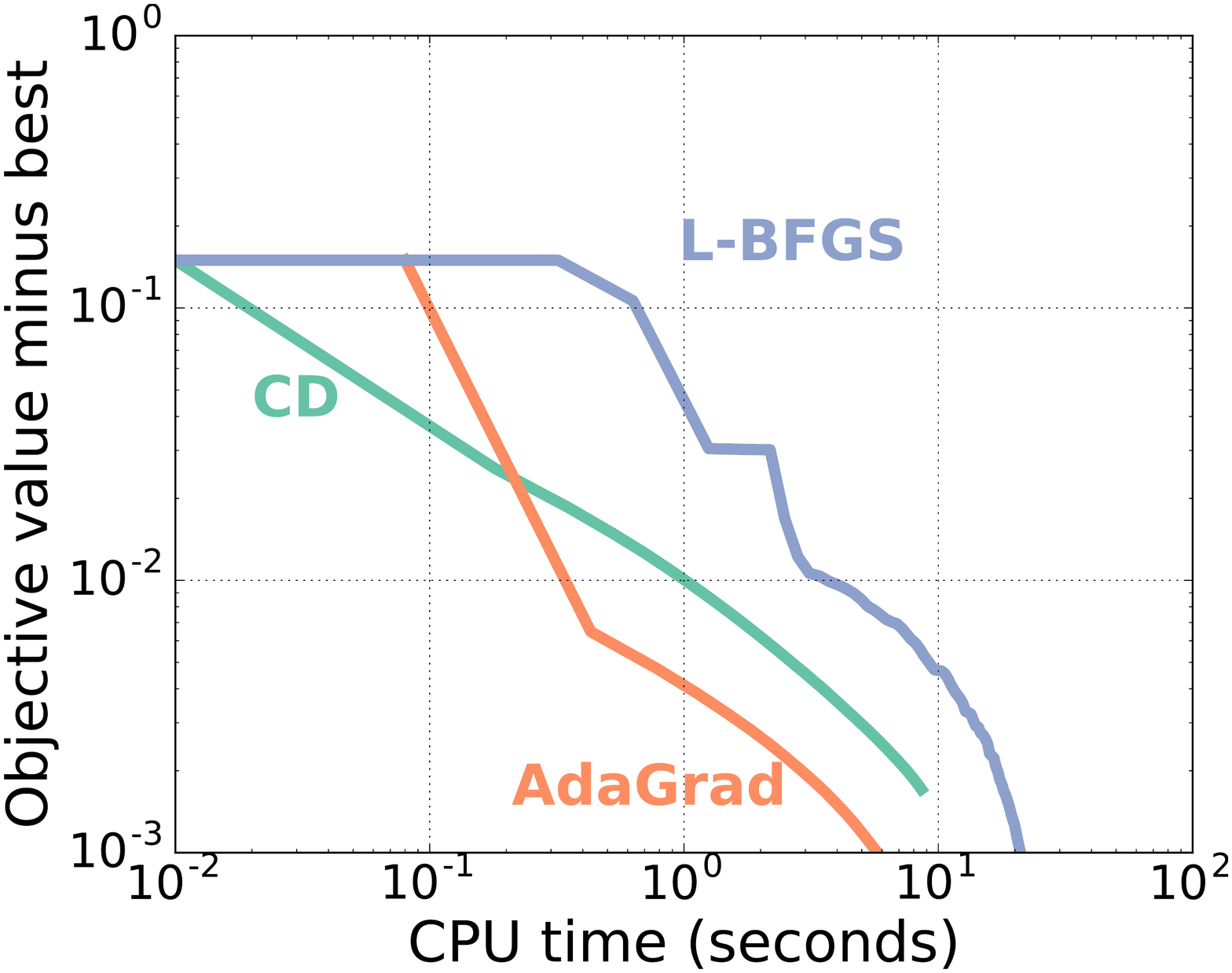}
}
\subfigure[$m=3$]{
    \includegraphics[scale=0.14]{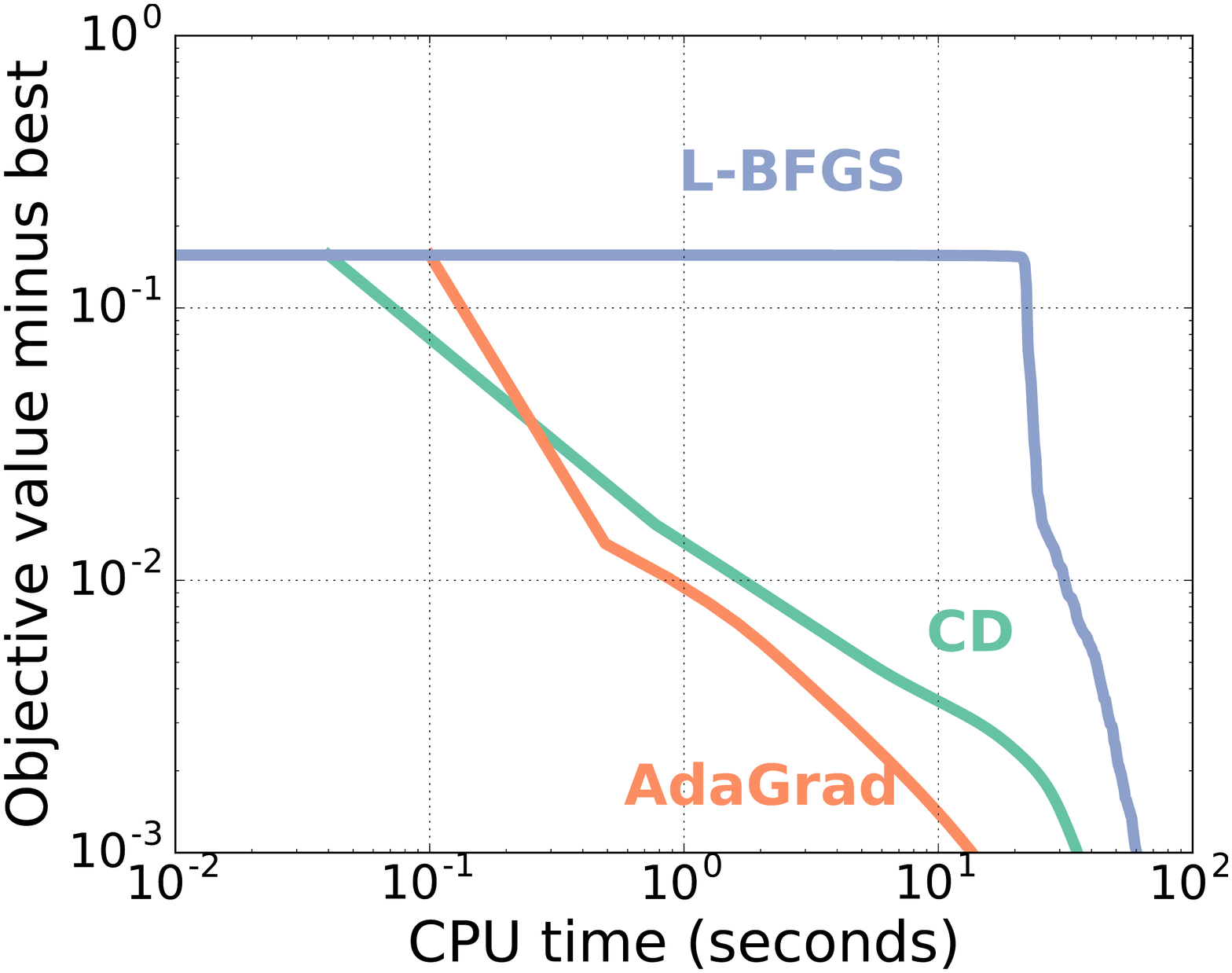}
}
\subfigure[$m=4$]{
    \includegraphics[scale=0.14]{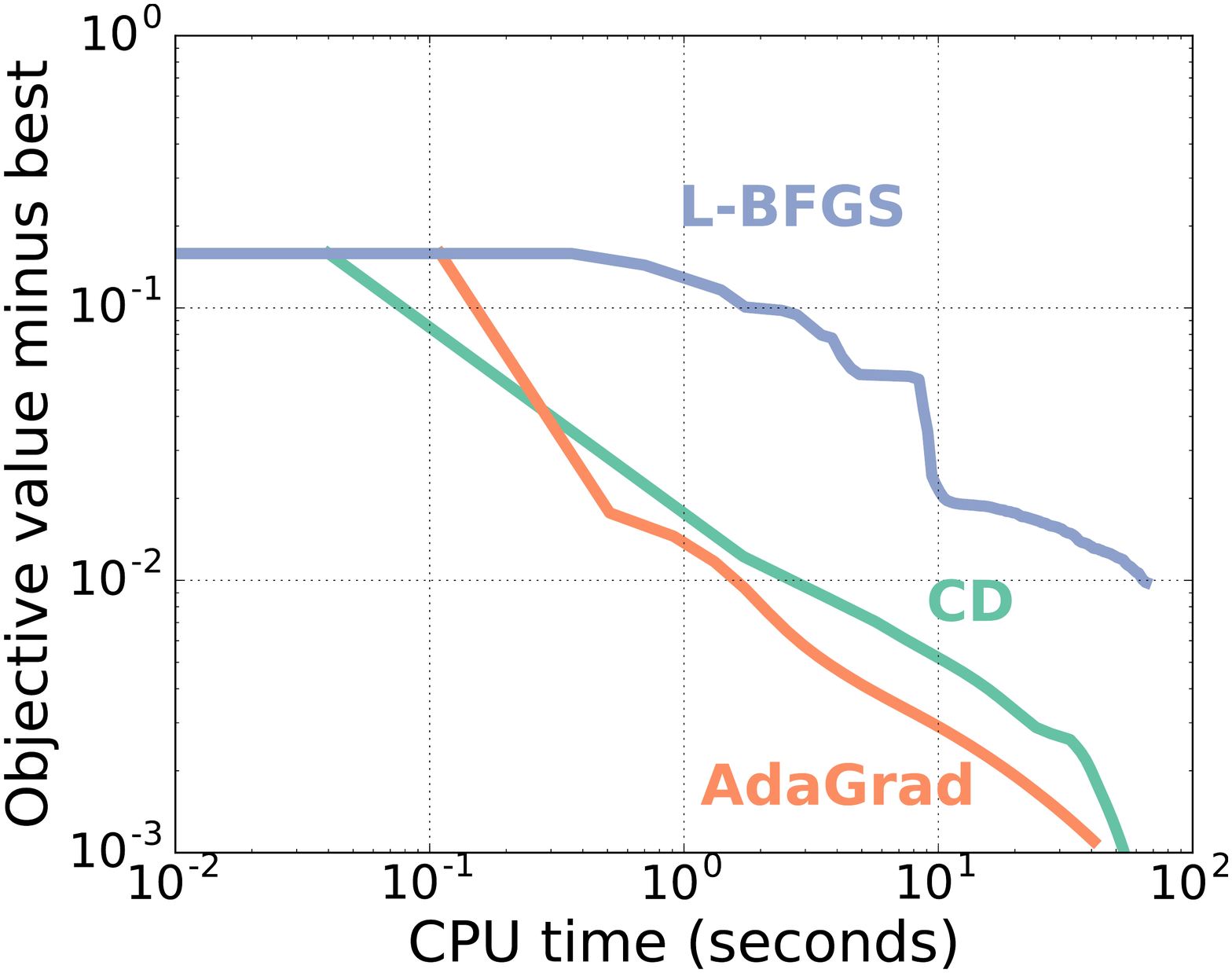}
}
\caption{Solver comparison for minimizing \protect\eqref{eq:obj_anova} on the
    Movielens 100K dataset. We set $\beta$ to the values with best test-set
    performance, which were $\beta=10^{-3}$, $\beta=10^{-4}$ and
    $\beta=10^{-6}$, respectively. We set $k=30$.}
\label{figure:solver_cmp4}
\end{figure}

We also compared AdaGrad, L-BFGS and coordinate descent (CD) on the Enzyme,
Gene-Disease (GD) and Movielens 100K datasets. Results are indicated in Figure
\ref{figure:solver_cmp2}, \ref{figure:solver_cmp3} and \ref{figure:solver_cmp4},
respectively.

\subsection{Recommender system experiments}

As we explained in Section \ref{sec:all_subsets}, the all-subsets kernel can be
a good choice if the number of non-zero elements per sample is small. To verify
this assumption, we ran experiments on two recommender system tasks: Movielens
1M and Last.fm. We used the exact same setting as in \cite[Section
9.3]{fm_icml}.  For each rating $y_i$, the corresponding $\bs{x}_i$ was set to
the concatenation of the one-hot encodings of the user and item indices. 
We compared the following models:
\begin{itemize}
    \item FM: $\hat{y}_i = \langle \bs{w}, \bs{x}_i \rangle + \sum_{s=1}^k
        \mathcal{A}^2(\bs{p}_s, \bs{x}_i)$
    \item FM-augmented: $\hat{y}_i = \sum_{s=1}^k \mathcal{A}^2(\bs{p}_s,
        \bs{\tilde{x}}_i)$ where $\bs{\tilde{x}}_i^\tr = [1, \bs{x}_i^\tr]$
    \item All-subsets: $\hat{y}_i = \sum_{s=1}^k \mathcal{S}(\bs{p}_s,
        \bs{x}_i)$ 
    \item Polynomial networks: $\hat{y}_i = \bs{\tilde{x}}_i \bs{U} \bs{V}^\tr
        \bs{\tilde{x}}_i$ (c.f. \cite{fm_icml} for more details)
\end{itemize}

\begin{figure}[p]
\center
\subfigure[Movielens 1M]{
    \includegraphics[scale=0.22]{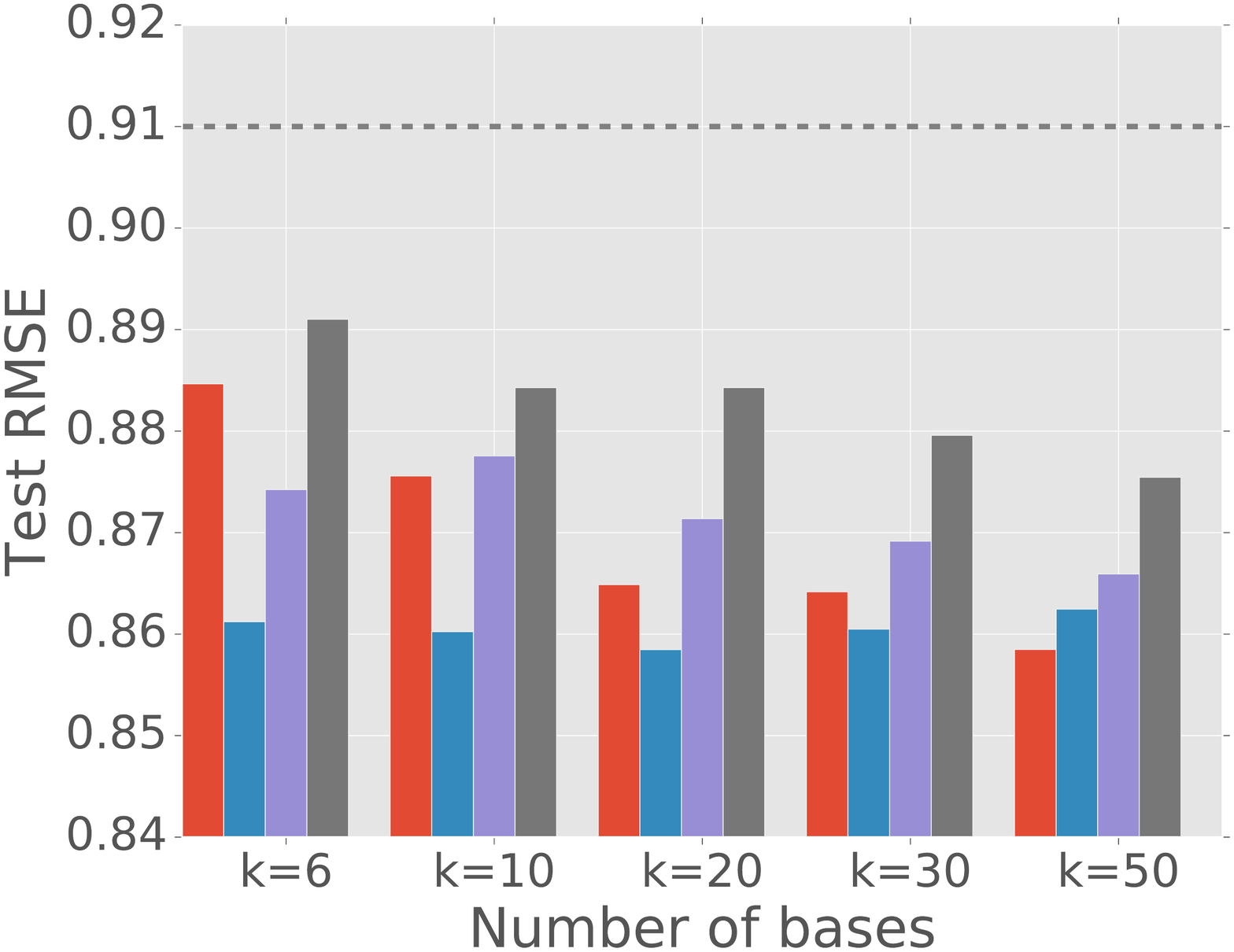}
}
\subfigure[Last.FM]{
    \includegraphics[scale=0.22]{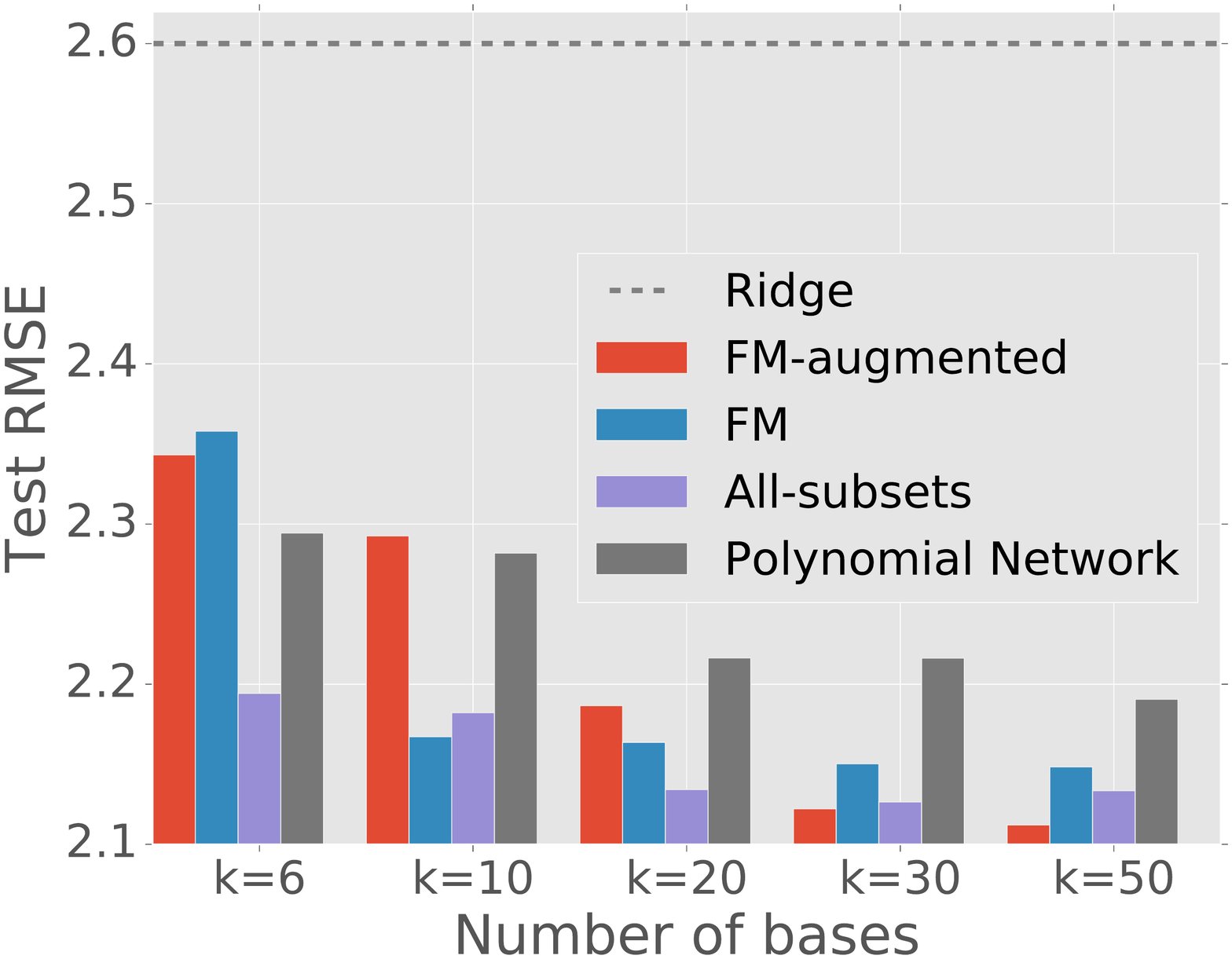}
}
\caption{Model comparison on two recommender system datasets.}
\label{figure:recsys}
\end{figure}

Results are indicated in Figure \ref{figure:recsys}. We see that All-subsets
performs relatively well on these tasks.

\clearpage
\section{Reverse-mode differentiation on the alternative recursion}
\label{appendix:alt_dp_gradient}

We now describe how to apply reverse-mode differentiation to the alternative
recursion \eqref{eq:recursion2} in order to compute the entire gradient
efficiently. Let us introduce the shorthands $a_t \coloneqq
\anovakern{p}{x}{t}$ and $d_t \coloneqq \diagkern{p}{x}{t}$.
We can then write the recursion as
\begin{equation}
    a_m = \frac{1}{m} \sum_{t=1}^m (-1)^{t+1} a_{m-t} d_t.
\label{eq:recursion2_shorthand}
\end{equation}
For concreteness, let us illustrate the recursion for $m=3$. We have
\begin{equation}
    a_1 = a_0 d_1, \quad
    a_2 = \frac{1}{2} (a_1 d_1 - a_0 d_2) \quad \text{and} \quad
    a_3 = \frac{1}{3} (a_2 d_1 - a_1 d_2 + a_0 d_3).
\end{equation}
We see that $a_2$ influences $a_3$, and $a_1$ influences $a_2$ and $a_3$.
Likewise, $d_3$ influences $a_3$, $d_2$ influences $a_2$ and $a_3$, and $d_1$
influences $a_1$, $a_2$ and $a_3$. Let us denote the adjoints $\tilde{a}_t
\coloneqq \partialfrac{a_m}{a_t}$ and $\tilde{d}_t \coloneqq
\partialfrac{a_m}{d_t}$. For general $m$, summing over quantities that
influences $a_t$ and $d_t$, we obtain
\begin{equation}
    \tilde{a}_t = \sum_{s=t+1}^m \frac{(-1)^{s-t+1}}{s} \tilde{a}_s d_{s-t}
    \quad \text{and} \quad
    \tilde{d}_t = (-1)^{t+1} \sum_{s=t}^m \frac{1}{s} \tilde{a}_s a_{s-t}.
\end{equation}
Let us denote the adjoint of $p_j$ by $\tilde{p}_j \coloneqq
\partialfrac{a_m}{p_j}$.  We know that $p_j$ directly influences only $d_1,
\dots, d_m$ and therefore
\begin{equation}
\tilde{p}_j = \sum_{t=1}^m \partialfrac{a_m}{d_m} \partialfrac{d_m}{p_j}
= \sum_{t=1}^m \tilde{d}_t t p_j^{t-1} x_j^t.
\end{equation}
Assuming that $d_1,\dots,d_m$ and $a_1,\dots,a_m$ have been previously computed,
which takes $O(dm+m^2)$, the procedure for computing the gradient can be
summarized as follows:
\begin{enumerate}
    \item Initialize $\tilde{a}_m = 1$,
    \item Compute $\tilde{a}_{m-1}, \dots, \tilde{a}_1$ (in that order),
    \item Compute $\tilde{d}_{m}, \dots, \tilde{d}_1$,
    \item Compute $\nabla \anovakern{p}{x}{m} = [\tilde{p}_1, \dots,
        \tilde{p}_d]^\tr$.
\end{enumerate}
Steps 2 and 4 both take $O(m^2)$ and step 4 takes $O(dm)$ so the total cost is
$O(dm+m^2)$. We can improve the complexity of step 4 as follows. We can rewrite
$\nabla \anovakern{p}{x}{m}$ in matrix notation:
\begin{equation}
\nabla \anovakern{p}{x}{m} = 
\left(
\begin{bmatrix}
1 & p_1 x_1 & (p_1 x_1)^2 & \dots & (p_1 x_1)^{m-1}\\
1 & p_2 x_2 & (p_2 x_2)^2 & \dots & (p_2 x_2)^{m-1}\\
\vdots & \vdots & \vdots & \ddots &\vdots \\
1 & p_d x_d & (p_d x_d)^2 & \dots & (p_d x_d)^{m-1}
\end{bmatrix}
\begin{bmatrix}
\tilde{d}_1 \\
2 \tilde{d}_2 \\
\vdots \\
m \tilde{d}_m
\end{bmatrix}
\right)
\circ \bs{x}.
\end{equation}
The left matrix is called a Vandermonde matrix. The
product between a $d \times m$ Vandermonde matrix and a $m$-dimensional vector
can be computed using the Moenck-Borodin algorithm (an algorithm similar to the
FFT), in $O(r \log^2 l)$, where $r=\max(d,m)$ and $l=\min(d,m)$
\cite{poly_book}. Since $m \le d$, the cost of step 4 can therefore be reduced
to $O(d \log^2 m)$.

\end{document}